\documentclass[journal]{IEEEtran}
\usepackage{amsmath,amsfonts}
\usepackage{algorithmic}
\usepackage{algorithm}
\usepackage{array}
\usepackage[caption=false]{subfig}
\usepackage{textcomp}
\usepackage{stfloats}
\usepackage{hhline}
\usepackage{soul,color}
\usepackage{multirow}
\usepackage{amsfonts}
\usepackage{float}
\usepackage{amsmath}
\usepackage{boldline} 
\usepackage{adjustbox}
\usepackage{xcolor}
\usepackage{arydshln}

\usepackage{url}
\usepackage{verbatim}
\usepackage{graphicx}
\usepackage{cite}
\usepackage{kotex}
\usepackage{kotex-logo}
\usepackage{amssymb}
\usepackage{pifont}
\newcommand{\cmark}{\ding{51}}%
\newcommand{\xmark}{\ding{55}}%

\usepackage[
  separate-uncertainty = true,
  multi-part-units = repeat
]{siunitx}

\begin{document}

\title{
Goal-conditioned dual-action imitation learning for dexterous dual-arm robot manipulation
}

\author{Heecheol Kim$^{1}$$^{,}$$^{2}$, Yoshiyuki Ohmura$^{1}$, Yasuo Kuniyoshi$^{1}$%
\thanks{$^{1}$ Laboratory for Intelligent Systems and Informatics, Graduate School of Information Science and Technology, The University of Tokyo, 7-3-1 Hongo, Bunkyo-ku, Tokyo, Japan (e-mail: \{h-kim, ohmura, kuniyosh\}@isi.imi.i.u-tokyo.ac.jp, Fax: +81-3-5841-6314) }%
\thanks{$^{2}$ Corresponding author}
}



\maketitle

\begin{abstract}
Long-horizon dexterous robot manipulation of deformable objects, such as banana peeling, is a problematic task because of the difficulties in object modeling and a lack of knowledge about stable and dexterous manipulation skills. This paper presents a goal-conditioned dual-action (GC-DA) deep imitation learning (DIL) approach that can learn dexterous manipulation skills using human demonstration data. Previous DIL methods map the current sensory input and reactive action, which often fails because of compounding errors in imitation learning caused by the recurrent computation of actions. The method predicts reactive action only when the precise manipulation of the target object is required (local action) and generates the entire trajectory when precise manipulation is not required (global action). This dual-action formulation effectively prevents compounding error in the imitation learning using the trajectory-based global action while responding to unexpected changes in the target object during the reactive local action. The proposed method was tested in a real dual-arm robot and successfully accomplished the banana-peeling task.

\end{abstract}


\section{Introduction}
\begin{figure}[!t]
\centering
\subfloat[Part of the automated banana peeling sequence in which the robot extends its end-effector to the banana  ($1$--$2$) and peels it ($3$--$4$). ]{\includegraphics[width=0.9\linewidth]{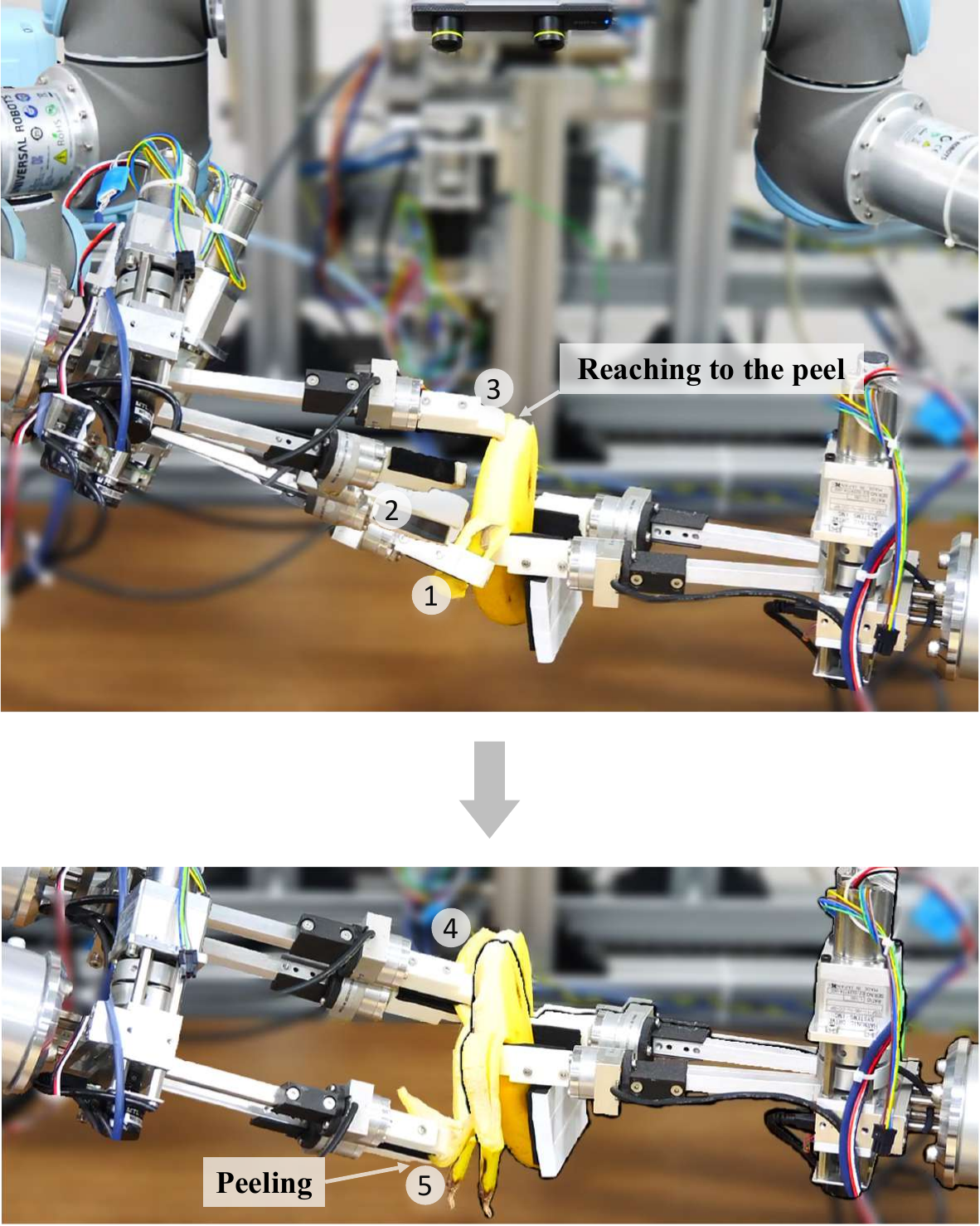}
\label{fig:sample_peel}}
\hfil
\subfloat[Example bananas used in the evaluation.]{\includegraphics[width=0.98\linewidth]{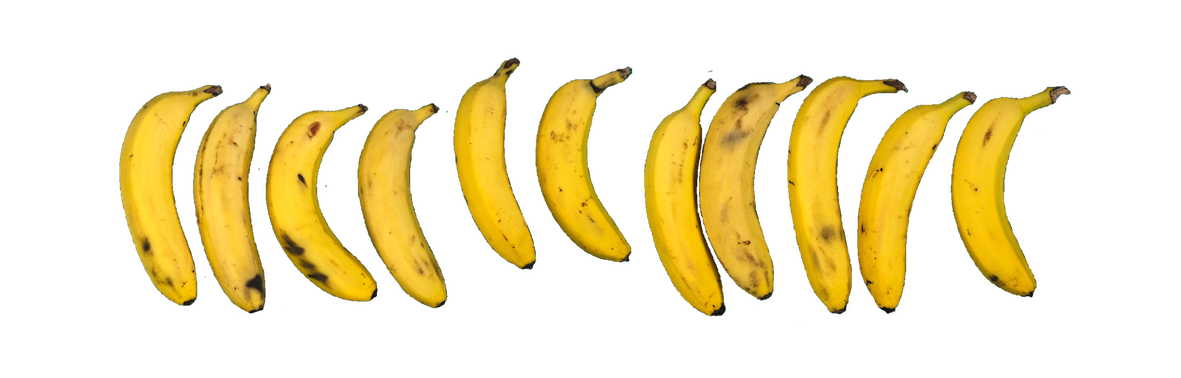}
\label{fig:sample_banana}}
\caption{Proposed goal-conditioned dual-action \textit{(GC-DA)} imitation learning (a) can peel real bananas of various shapes (b) using a general-purpose dual-arm robot.
}\label{fig:sample_first}
\end{figure}

\IEEEPARstart{D}{eep} imitation learning, which trains the robot's behavior using human-generated demonstration data with deep neural networks, is a promising technique because it can transfer the implicit human knowledge of dexterous manipulation into a robot system without a pre-defined manipulation rule based on knowledge of the objects \cite{finn2016deep,zhang2018deep,kim2021gaze,kim2021transformer}. 

This paper presents a deep imitation learning approach to achieve dexterous robot manipulation tasks such as banana peeling (Fig. \ref{fig:sample_first}) with a general-purpose dual-arm robot manipulator (UR5, Universal Robots). In this task, every banana's shape, size, and skin pattern are different (Fig. \ref{fig:sample_banana}), making banana model acquisition difficult. In addition, the deformation of the banana during peeling makes this task even more difficult. 
The diversity in the sensory input state, brought about by the varying nature of the banana, necessitates high precision in the robot's actions to ensure the policy remains within the permitted action space. For example,
because the banana is deformable, soft, and easily breakable, the robot has to accurately position its finger on the banana's peel and tear it away while maintaining its body; this requires highly dexterous and precise manipulation skills that consider physical interactions with the target object, which are difficult for rule-based robotics to acquire. Moreover, the banana peeling is a difficult task that consists of a combination of several reaching and peeling subtasks and requires stable manipulation policies for consecutive success on each subtask. Therefore, to our knowledge, no previous robotics research has tackled this task.

Our previous research proposed a visual attention-based deep imitation learning framework \cite{kim2020using} that can focus on an important area of visual input by using the gaze signal of a human expert while demonstrating the task. Subsequently, \cite{kim2021gaze} proposed a dual-resolution dual-action system that handles local action in the vicinity of the gaze position separately from global action outside of the gazed area. This approach has successfully completed dexterous manipulation tasks such as needle threading. However, the dual-resolution dual-action system was applied to tasks in which the objects are located in two-dimensional space, which is narrower than the banana-peeling task. Moreover, the state variation of the needle is narrower because the needle is not deformable. 
Therefore, the previous dual-resolution dual-action system cannot provide data efficiency and a stable trajectory for the banana-peeling task.

\begin{figure*}[!t]
\centering
\includegraphics[width=0.98\linewidth]{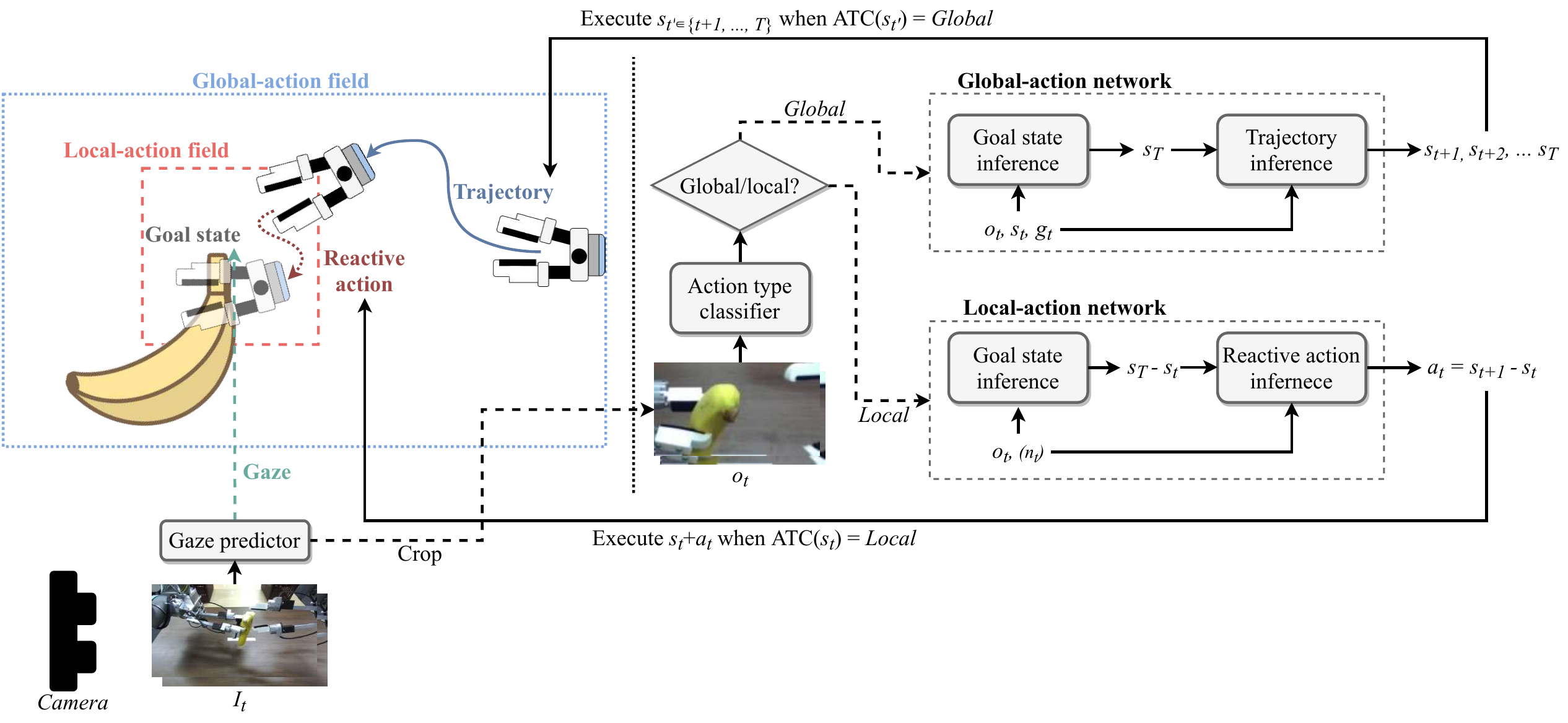}
\caption{Proposed \textit{GC-DA} architecture. The action-type classifier (ATC) determines if precise manipulation of the target object is required; if not, the global-action network computes the trajectory of the end-effector, which is robust against compounding errors; otherwise, the local-action network computes a reactive action that can dexterously manipulate the target object. In practice, once the ATC's output transitions from local-action to global-action or global-action to local-action, the ATC's output is fixed, improving system robustness despite possibly noisy output of the ATC. Here, $o_t$, $n_t$, $s_t$, $g_t$, $a_t$, and $T$ refer to the foveated image, force/torque sensory values, robot kinematics states, gaze position, action, and last timestep in the sequence, respectively. Detailed neural network architectures for the global-action network, local-action network, and gaze-predictor can be found in Fig. \ref{fig:network}.}
\label{fig:concept}
\end{figure*}

The problems of the current dual-action system described above arise from the compounding error caused by reactive action. 
In the previous dual-action system, both global and local actions are a reactive action, which is the difference between the state in the next step and the current state, that is, $a_t = s_{t+1} - s_t$, where $a_t$ is the reactive action and $s_t$ is the robot kinematic state.
This reactive action can respond to unexpected immediate changes in the environment caused by an incomplete environment model and error in policy predictions. However, because the reactive action changes the robot state recurrently, in long-horizon tasks, the state eventually escapes from the training distribution because of compounding error in the action prediction. Moreover, if the agent predicts the entire trajectory, the problems caused by the compounding error occur less frequently, but the trajectory-based manipulation cannot deal with immediate changes in the environment.

We hypothesize that in the manipulation task, the immediate changes in the environment usually occur while the end-effector is interacting with the target object. Therefore, if the end-effector is in the vicinity of the object, reactive action is required; if it is not, trajectory-based action can stably transport the robot toward the goal position. 

On the basis of the hypothesis described above, we propose a method that uses reactive action for dexterous manipulation and trajectory-based action for robustness against compounding error (Fig. \ref{fig:concept}). This architecture separately handles local action in the sequence of the robot behavior, where the local action is defined as the reactive action output from the visually attended feature input images using robot gaze (foveated images). The global action is a trajectory $s_{t+1}, s_{t+2}, ..., s_{T}$, where $T$ indicates the last step in the sequence, and is computed from the foveated image, current robot kinematic state, and gaze position.
A simple CNN-based classifier can switch the action between global and local action based on a criterion that considers whether precise control of the target object is required.

This research also improves manipulation performance by predicting the goal state, which is defined as the robot's kinematic state at the last timestep of each subtask, and the goal-conditioned action inference. Previous deep imitation learning methods predicted reactive action from the input states of the current timestep \cite{finn2016deep,zhang2018deep,kim2020using}. However, action should also be conditioned by the goal state because the action can be considered a part of the route from the current state to the goal state; therefore, the action must be changed if the goal state is changed. A neural network only statistically maps the action from the current sensory input state without goal conditioning, which is fragile to distractions in the input state. 
Therefore, the proposed model first predicts the goal state, which is defined as the last state in the sequence, and then the action is predicted using the goal state and the current sensory input state to achieve robust policy prediction. 

The proposed goal-conditioned dual-action (\textit{GC-DA}) imitation learning model was tested with the banana-peeling task in a real dual-arm robot manipulator (UR5).

To summarize, our contributions are as follows:
\begin{itemize}
    \item A dual-action architecture that is a combination of reactive local action and trajectory-based global action to achieve dexterity while maintaining stability in the manipulation policy.
    \item Goal-based policy inference using the dual-action framework.
    \item Evaluation of the proposed deep imitation learning framework using the challenging banana-peeling task with a real-world dual-arm robot.
\end{itemize}

\section{Related work}
\subsection{Trajectory-based methods for robot manipulation}
Trajectory generation methods have been studied widely in robotics (e.g., \cite{koert2016demonstration,vakanski2012trajectory}). For example, \cite{kim2020reinforcement} studied searching trajectories in reinforcement learning using a factorized-action variational autoencoder \cite{yamada2020disentangled}. Notably, in path planning methods in which a robot plans its path to a goal while avoiding obstacles, global path planning and local path planning are present. Global path planning plans the entire path using global information \cite{turpin2014goal,bhattacharya2015persistent,van2014optimal,gammell2015batch}. By contrast, the local path planning computes real-time reactive action based on local information and is robust to unpredicted disturbances in the environment \cite{khatib1986real,hwang1992potential,sedighi2004autonomous,cole2017reactive}. Hierarchical planners that combine global and local path planning have achieved collision-free motion without delays caused by the local path planner \cite{knepper2010hierarchical}. This approach supports our system, which combines trajectory-based global action and reactive local action. Nevertheless, these path planning-based methods cannot be directly applied to our target task because acquiring a deformable and high-variance object model from visual sensory input is difficult, and both the path planning and reactive action with the target object are unknown. 

\subsection{Goal-based policy learning}
Many different aspects of goal-based policy learning have been studied. \cite{andrychowicz2017hindsight,ding2019goal,lee2021generalizable} considered a goal while searching for a policy during manipulation. \cite{andrychowicz2017hindsight} proposed hindsight experience replay (HER), which assumes the last state in an explored episode is the goal, even if the episode was not successful. This goal is used for the data-efficient training of a goal-conditioned policy. \cite{ding2019goal} applied HER to generative adversarial imitation learning \cite{ho2016generative}, which uses generative adversarial training \cite{goodfellow2014generative} for reward learning to leverage policy convergence speed. \cite{lee2021generalizable} acquired dense rewards using a goal proximity function trained using demonstration data. However, exploration-based methods cannot be applied to our task setup, which must train a dexterous robot policy using only a limited number of bananas as training data.

Several studies do not rely on exploration in addition to demonstration data \cite{sui2017goal,zeng2018semantic,lynch2019learning,codevilla2018end,friesen2010imitation,dindo2010adaptive,mandlekar2020iris,mandlekar2020learning}. \cite{sui2017goal,zeng2018semantic,lynch2019learning,codevilla2018end,mandlekar2020iris,mandlekar2020learning} employed the goal to help predict the policy. \cite{sui2017goal,zeng2018semantic} used user-provided goal scenes to select pre-defined manipulation behavior from an axiomatic scene graph of current visual input. These methods successfully manipulated objects in long-horizon tasks in cluttered scenes. \cite{friesen2010imitation,dindo2010adaptive} selected high-level action using goal-based imitation, where the low-level actions are pre-defined. \cite{lynch2019learning,codevilla2018end} inferred the primitive action from the goal. \cite{lynch2019learning} trained a latent variable model using a variational autoencoder \cite{kingma2013auto} to control a robot manipulator toward the given goal. \cite{codevilla2018end} proposed a neural network architecture conditioned on high-level commands for autonomous driving. 
In these methods, the goal is used as a signal to change the agent's behavior. By contrast, in our proposed method, the goal prediction and conditioning are employed to achieve a stable and dexterous manipulation policy.

\cite{zhang2018deep} improved manipulation accuracy by training the auxiliary prediction, which estimates the gripper's position when closed during picking. \cite{mandlekar2020iris} leveraged goals to manage the high variability in demonstration quality and task solution approaches commonly seen in human expert data. \cite{mandlekar2020learning} utilized the intersecting structure of the training dataset to formulate a policy that can generalize to unseen start and goal state combinations. In contrast, our goal-conditioned approach aims to generate a stable policy for the highly complex task of banana peeling. Our demonstration set, derived from a highly skilled expert, inherently exhibits low variability in subgoals and there are few alternative policies for achieving the subgoals.


\subsection{Automation in food processing}
Studies on food processing automation using robots include food manufacturing in factories \cite{caldwell2009automation,wilson2010developments}, fruit harvesting \cite{wang2019window,zhang2019multi,yu2019fruit,kuznetsova2020using,huangfei2019design,mu2020design}, and cooking \cite{bollini2013interpreting,zhang2019robot}. Many of these studies relied on task-specific end-effector designs \cite{davis2007robot,ma2020made,huangfei2019design,mu2020design} or rule-based manipulation \cite{wang2019window,zhang2019multi,yu2019fruit}. 
Therefore, they lacked the acquisition of dexterous manipulation skills, instead, they limited the target task to simple object manipulations, such as pick-and-place tasks. 

Some work has studied long-horizon deformable object (dough) manipulation in a simulator \cite{lin2022diffskill} and real-world \cite{lin2022planning}. However, learning for the banana-peeling task requires highly precise manipulation skills in three-dimensional state space, which requires skills that are substantially different to those required for dough control.

\cite{lee2020guided,yamada2021motion} combined an open-loop motion planner and closed-loop neural-network-based policy, leveraging the merits of both methods. However, these methods require a reinforcement-learning-based policy search, which is infeasible in deformable and destructible object manipulation and requires an additional pipeline with an open-loop motion planner. By contrast, the proposed \textit{GC-DA} relies on the imitation learning framework, which is much more sample efficient and requires no additional traditional robot control pipeline but can still leverage the strength of the combined robust open-loop control and precise closed-loop object manipulation.

Some learning-based methods have been applied to food processing as target tasks, such as pouring \cite{liu2018imitation,smith2019avid} or stirring \cite{yang2022viptl}. However, those tasks usually do not require manipulation skills as precise as those needed in our target task of banana peeling; therefore, a new method is needed for learning highly skilled tasks.

Hence, this research uses a general-purpose manipulator with a single degree-of-freedom end-effector, which can also be applied to other various tasks \cite{kim2020using,kim2021gaze,kim2021transformer,kim2022training}, to learn dexterous manipulation skills such as banana-peeling by imitating human manipulation skills. 

\section{Preliminaries}
\subsection{Gaze-based attention}
First proposed in \cite{kim2020using}, gaze-based attention is a method that acquires visual attention using a human's gaze information, obtained during demonstrations. This method measures the human gaze during a demonstration with an eye tracker and trains the neural network model to output the gaze's probability distribution from the camera's visual input. Using the model, the robot can visually attend to the most probable position in a given visual scene. Using the predicted gaze coordinates, the model crops the foveated vision from the entire raw pixel image, and this foveated vision is used as the visual input to another neural network model, which computes the action policy. An important advantage of using gaze-based attention is that it significantly reduces the number of network parameters and the size of the training dataset for tasks that necessitate high-resolution vision for precise manipulation \cite{kim2021gaze}. Without the gaze model, the neural network would be obligated to process the entire high-resolution vision (1280 x 720 for this task). However, the majority of these pixels contain irrelevant information, such as the background and unrelated objects. This leads to impractical training time and space requirements. Also, Because the task-unrelated background and other objects are masked out in the foveated vision, gaze-based attention is robust against the distractions in the background and task-unrelated objects and is more computationally efficient \cite{kim2020using}. Fig. \ref{fig:network} (c) illustrates this model.

\subsection{Dual-action}\label{dual-action}
First proposed in our previous study \cite{kim2021gaze}, dual-action is a method that distinguishes actions in the global reaching behavior and precise object manipulation to enable dexterous manipulation skills while maintaining global reaching ability. This method handles the local action, which consists of precise manipulation skills that require interaction with the target object, separately from the global action, which delivers the robot hand to the area near the target. The local-action network is trained only with local action so that it is not affected by disturbances from the global action. In our previous study \cite{kim2021gaze}, both the global and local actions are reactive and not conditioned on the goal state. In these studies, the criterion for distinguishing global/local action was based on speed, but this does not apply to general tasks, where a significant difference in speed between local and global actions is not guaranteed. Therefore, we annotated the global/local action manually based on whether the robot's end-effector is in the foveated vision.

\section{Method}

\subsection{Robot framework}
This research uses a dual-arm robot framework for imitation learning that was also used in our previous studies (e.g., \cite{kim2020using,kim2021gaze}). This framework consists of a dual-arm robot system with two UR5 manipulators and two controllers with identical kinematic parameters with the UR5 robot. Demonstration data are generated by controlling the robots with the controllers. 
A ZED mini stereo camera (StereoLabs, \cite{ZED-MINI}) is mounted on the robot system with a two-dimensional pan-tilt structure. In this research, the camera is fixed at a position that can observe bananas (pan: 0 rad, tilt: -0.9 rad). The human operator can see the stereo camera image through a head-mounted display while operating the robot. Furthermore, an eye tracker (Tobii) is mounted on the head-mounted display to measure the operator's gaze position in real time.

In this method, $s_t$ is the robot state at timestep $t$ defined by concatenated left and right arm states. Each arm state consists of three dimensions to express the position (x,y,z), six dimensions for the cosine and sine of the Euler angle for rotation, and one dimension for the gripper angle. The Euler angle is decomposed into cosine and sine expressions to enable a smooth expression when the angle exceeds $2*\pi$.
The action of each arm is hence a seven-dimensional vector with the differences in position (three dimensions), Euler angle (three dimensions), and gripper (one dimension) in adjacent timesteps.

\subsection{Task specification}\label{subsection:task_specification}
The banana-peeling task can be segmented into several subtasks (Table \ref{tab:subtask_details}).
The entire banana-peeling sequence is segmented using natural semantic criteria: the robot first grasps the banana 
(\textit{GraspBanana}), then picks the banana up (\textit{PickUp}), moves its arm to the banana's neck (\textit{GraspNeck}), peels it (\textit{PeelNeck}), moves its arm to the banana's peel (\textit{ReachRight}), peels it (\textit{PeelRight}), rotates the banana so that the right hand can reach the left peel (\textit{Reposition}), moves the arm to the banana's last peel (\textit{ReachLeft}), and then peels it (\textit{PeelLeft}). 

\begin{figure*}[!t]
\centering
\includegraphics[width=0.7\linewidth]{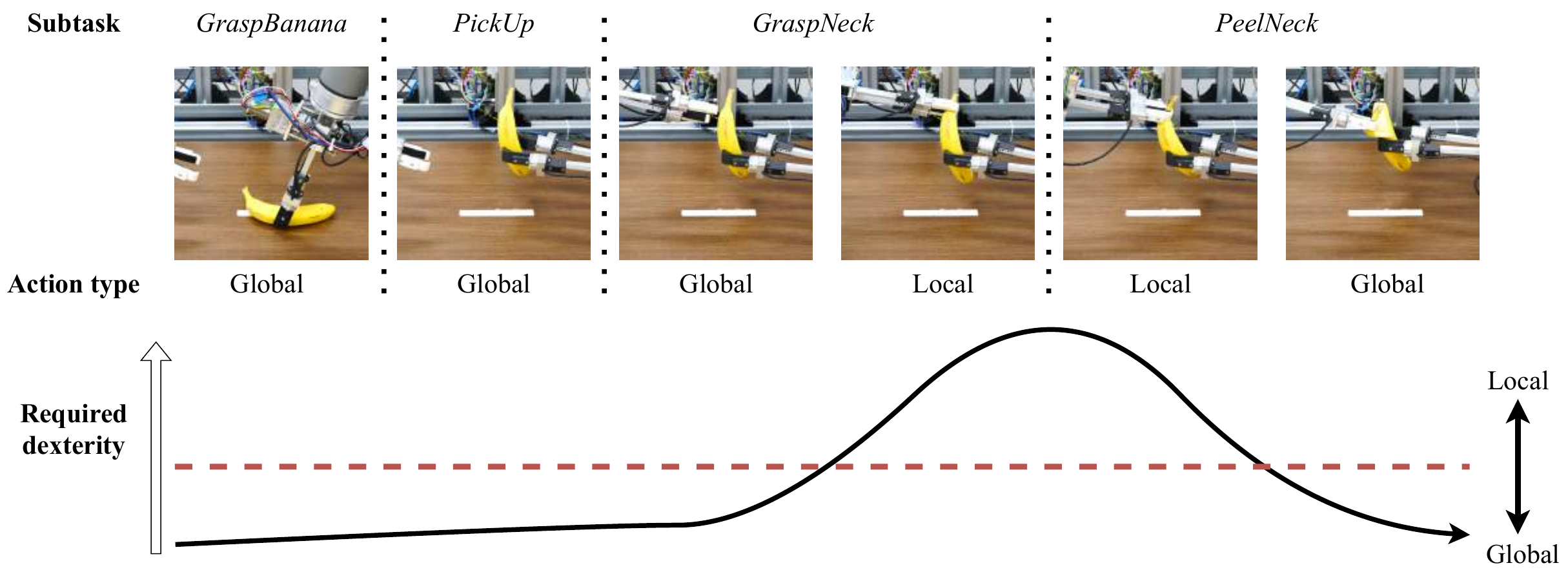}
\caption{
Relationship between the required dexterity for the subtasks and connection to the action type. Local action is used for the tasks that require more dexterity.
}
\label{fig:task_desc}
\end{figure*}


Each subtask consists of a combination of global and local actions. Among all possible combinations of dual-action (global-only, local-only, global-to-local, and local-to-global), the subtask selects one based on the required dexterity. As shown in Fig. \ref{fig:task_desc}, each subtask requires different levels of dexterity. For example, \textit{GraspNeck, ReachRight}, and \textit{ReachLeft} do not require high dexterity while moving the robot hand to the vicinity of the neck or peels, but skillful manipulation is required when the hand is near the target and tries to reach the exact position. During peeling subtasks, a closed-loop control is needed until the robot has peeled the target. Once the target has been peeled, less dexterity is required. Therefore, the global-to-local combination is used for grasping and reaching subtasks, and peeling subtasks use local-to-global combinations. Simple subtasks that do not require dexterity  (\textit{GraspBanana, PickUp}, and \textit{Reposition}) {use global-only action.}

A next step classifier (NSC) is a neural network that detects whether the current executing subtask needs to end and to robot should transit to the next subtask. This network is required for subtasks that end with a local action 
(i.e., \textit{GraspNeck} to \textit{PeelNeck}, \textit{ReachRight} to \textit{PeelRight}, and \textit{ReachLeft} to \textit{PeelLeft}). 
NSC is composed of a multi-layer perceptron (MLP) architecture followed by a Conv module (Fig. \ref{fig:network}). It uses the foveated vision as input at outputs a binary classification result, where the former subtasks
(\textit{GraspNeck}, \textit{ReachRight}, and \textit{ReachLeft})
are assigned the label 0 and the latter subtasks (\textit{PeelNeck}, \textit{PeelRight}, and \textit{PeelLeft}) are assigned the label 1.

Another possible strategy for banana peeling is to re-grasp the banana with the right hand and peel the left side with the left hand. However, the banana may become unstable after the re-grasping. Therefore, the banana is rotated rather than re-grasped. In addition, many people peel the banana from the banana's tip. However, the robot was not able to peel from the tip even with the human operator because of a lack of grasping power.

\begin{table*}
\centering
\caption{Subtask details and a visual example of the goal. ``ATC'' indicates whether an action-type classifier (ATC) is required to detect the transition from a global to local (or local to global) action. ``NSC'' indicates whether the next step classifier (NSC) is required to detect the end of the subtask so that the next subtask can be executed. All subtasks end with local actions that require the NSC. An asterisk is used for consecutive peeling setups. Every subtask is trained with a separate model; subtasks do not share model parameters.}
\begin{tabular}{clcccc|clcccc}
\hlineB{2}
    Index & Subtask                   & Goal & Network type & ATC & NSC
    & Index & Subtask & Goal & Network type & ATC & NSC \\ \hline \hline
1 &\textit{GraspBanana}                  & \adjustbox{valign=c}{\includegraphics[width=1.8cm, height=1.8cm]{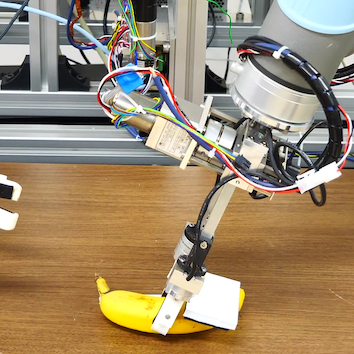}}
     & Global & \xmark & \xmark &
6 & \textit{ReachRight} &
\adjustbox{valign=c}{\includegraphics[width=1.8cm, height=1.8cm]{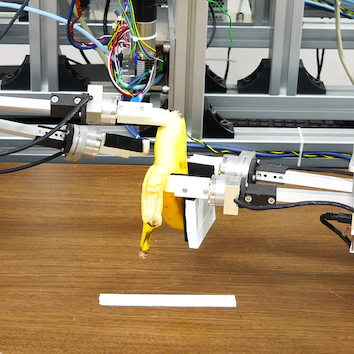}}& Global $\rightarrow$ Local & \cmark & \cmark\\ \hline

2 & \textit{PickUp} & 
\adjustbox{valign=c}{\includegraphics[width=1.8cm, height=1.8cm]{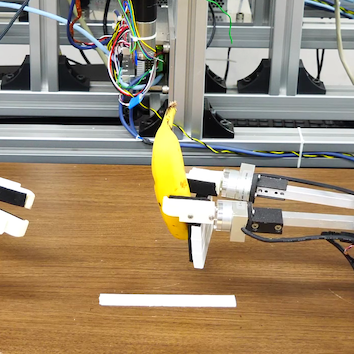}}& Global &  \xmark & \xmark &
7 & \textit{PeelRight} &
\adjustbox{valign=c}{\includegraphics[width=1.8cm, height=1.8cm]{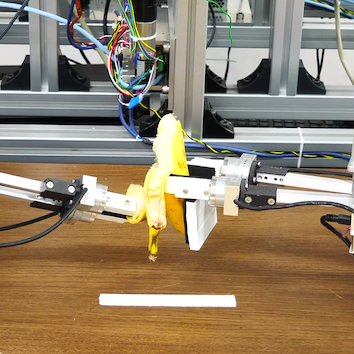}}& Local $\rightarrow$ Global & \cmark & \xmark \\ \hline

3 & \textit{GraspNeck} &
\adjustbox{valign=c}{\includegraphics[width=1.8cm, height=1.8cm]{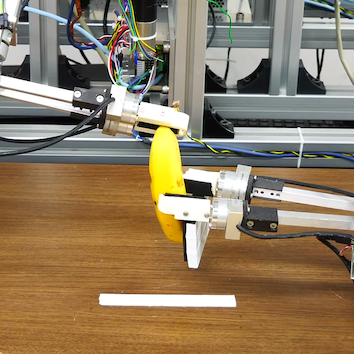}}& Global $\rightarrow$ Local & \cmark & \xmark & 
8 & \textit{Reposition} &
\adjustbox{valign=c}{\includegraphics[width=1.8cm, height=1.8cm]{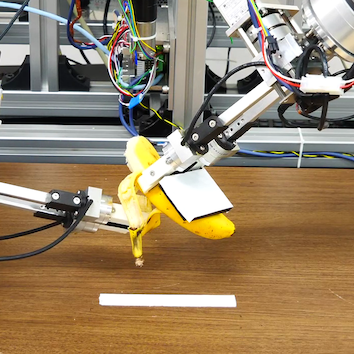}}& Global & \xmark & \xmark\\ \hline

4 & \textit{Push*} &
\adjustbox{valign=c}{\includegraphics[width=1.8cm, height=1.8cm]{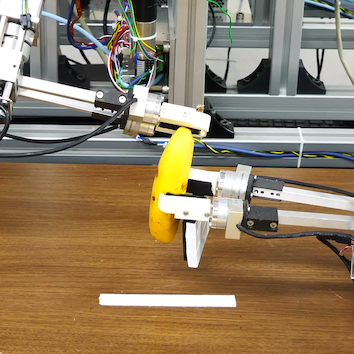}}& Hand-crafted & \xmark & \xmark &
9 & \textit{ReachLeft} &
\adjustbox{valign=c}{\includegraphics[width=1.8cm, height=1.8cm]{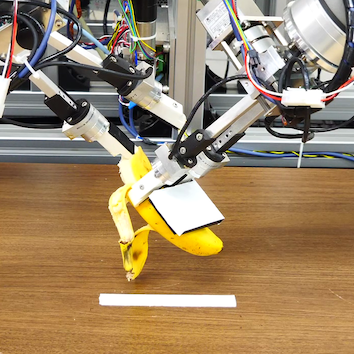}}& Global $\rightarrow$ Local & \cmark & \cmark \\ \hline

5 & \textit{PeelNeck} &
\adjustbox{valign=c}{\includegraphics[width=1.8cm, height=1.8cm]{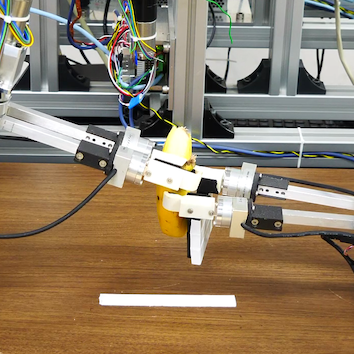}}& Local $\rightarrow$ Global & \cmark & \xmark &
10 & \textit{PeelLeft} &
\adjustbox{valign=c}{\includegraphics[width=1.8cm, height=1.8cm]{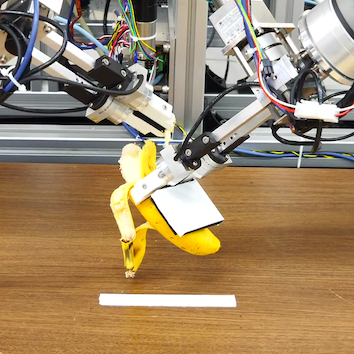}}& Local $\rightarrow$ Global & \cmark & \xmark \\ \hline
\hlineB{2}\\
\end{tabular}
\label{tab:subtask_details}
\end{table*}

During demonstrations, because the number of bananas was limited, non-destructive subtasks were repeated to maximize the number of demonstrations. \textit{GraspBanana}, \textit{PickUp}, \textit{GraspNeck}, \textit{ReachRight}, \textit{Reposition}, and \textit{ReachLeft} do not tear or break the target banana; therefore, those subtasks were repeated $10$--$20$ times per banana. By contrast, \textit{PeelNeck}, \textit{PeelRight}, and \textit{PeelLeft} peel the bananas; therefore, these subtasks are non-reversible. Thus, these subtasks were recorded only once per banana. All demonstrations were recorded at 5 Hz. 
Table \ref{tab:dataset_statistics} presents the statistics of each subtask.
\begin{table}
\centering
\begin{tabular}{lcc}
\hlineB{2}
    Dataset                   & Number of demos & Total demo time (min) \\ \hline \hline
\textit{GraspBanana}                  & 2050     & 89.9 \\
\textit{PickUp} & 2073     & 63.2 \\
\textit{GraspNeck} & 2141     & 80.9 \\ 
\textit{PeelNeck} & 256     & 35.6 \\ 
\textit{ReachRight} & 4458     & 211.6 \\ 
\textit{PeelRight} & 264     & 27.4 \\ 
\textit{Reposition} & 1779     & 41.3 \\ 
\textit{ReachLeft} & 4009     & 237.6 \\ 
\textit{PeelLeft} & 289     & 23.7 \\ \hline 
\textit{Total} & 17319 & 811.1 \\
\hlineB{2}\\

\end{tabular}
\caption{Training set statistics. }
\label{tab:dataset_statistics}
\end{table}


\subsection{Goal-conditioned dual-action}
Figure \ref{fig:concept} and Algorithm \ref{alg:algo} illustrate the concept of goal-conditioned dual action. The dual-action-based architecture separates the reactive local action, which is used when precise control is required, from the trajectory-based global action, which is used when path planning that is robust against compounding error is required. 

The boundary between local action and global action is decided by whether the robot requires precise manipulation. This boundary decision is annotated by humans and then trained using a binary classifier to autonomously switch between global and local action (see \ref{subsection:global_local_separation} for details). Notably, on high-precision \textit{GraspNeck/Reach}$\ast$ subtasks, a simple criterion, which is whether the end-effector is in the foveated image, can determine the boundary (see Appendix \ref{appendix:global_local} for details).

Both global and local actions are conditioned by the goal state, which is the robot kinematic state at the last timestep $T$ of the sequence in each subtask. Simple end-to-end imitation learning $a_t = s_{t+1} - s_t = \pi(e_t)$, where $\pi$, $e_t$, $s_t$, and $a_t$ respectively refer to the policy, embedding of the current input states, robot kinematic state, and current action, which maps $s_t$ to $a_t$, is unstable with respect to the distractions in $s_t$. 
By contrast, the explicit inferred goal state $s_T$, where $T$ refers the last timestep in the sequence, can provide a stable policy because the action can use explicit information about the task goal. 
The trajectory-based global-action network first predicts the goal state: $s_T = \pi^{goal}_{global}(e_t)$, where $s_T$ and $\pi^{goal}_{global}$ respectively refer to the inferred robot kinematic goal state and the goal predictor of the global-action network. Then, trajectory $s_{t+1}, s_{t+2}, ..., s_{T-1}$ is predicted using both $e_t$ and $s_T$:  $s_{t+1}, ..., s_{T-1} = \pi^{policy}_{global}(e_t, s_T)$, where $\pi^{policy}_{global}$ refers to the policy inference of the global-action network. For all subtasks except for \textit{GraspBanana}, \textit{PickUp}, and \textit{Reposition}, which are only composed of global action, the predicted trajectory is executed during $ATC(o_{t' \in \{t+1, ..., T\}}) = global$, where $ATC$ indicates the action-type classifier.

The action type classifier (ATC) is trained to predict the type of global/local action from the foveated vision as input. The purpose of this network is to predict the transition from global action to local action while executing the policy networks in the real robot environment. The dual-action system establishes a link between foveated vision and local action, an idea grounded in the findings of physiological studies \cite{paillard1996fast}. Consequently, the visual presence of the robot's gripper in the foveated vision serves as a logical criterion for action type identification. This network utilizes the same Conv module architecture (Fig. \ref{fig:network} (d)) and an MLP with two-dimensional output for the global/local labels.

The reactive local-action network only inputs the foveated image $o_t$ because the reactive local action can be computed efficiently from the relation between the target object and the end-effector \cite{kim2021gaze}. 
In this architecture, information about the global coordinate is lost, and the neural network only focuses on the relationship between the target object and the end-effector. Therefore, this network predicts and uses the goal state in a modified form; the relative goal state $s^{local}_T = s_T - s_t = \pi^{goal}_{local}(o_t)$ is inferred and used to predict the reactive action, where $\pi^{goal}_{local}$ is the goal predictor of the local-action network. Therefore, the reactive action is conditioned by the current foveated visual information and relative goal state: $a_t = s_{t+1} - s_t = \pi^{policy}_{local}(o_t, s^{local}_T)$, where $\pi^{policy}_{local}$ refers to the policy inference of the local-action network.

\begin{algorithm}[tb]
\caption{Goal-conditioned dual-action deep imitation learning}
\label{alg:algo}
\textbf{Parameters}: 
Action-type classifier ATC, Global-action network $\pi_{global}$, Local-action network $\pi_{local}$, gaze predictor $\rho$
\begin{algorithmic}[1] 
\STATE $t \gets 0$ \COMMENT{initialize timestep}
\WHILE{$\neg$ $succeed$}
\STATE $I_t$ $\gets$ $1280 \times 720 \times 6$ raw stereo camera image
\STATE $s_t \gets$ left, right robot kinematics states
\STATE $i_t \gets$ resize$(I_t, (128, 72))$ \COMMENT{resized stereo global image}
\STATE $g_t \gets \rho(i_t)$  \COMMENT{predicted gaze position}
\STATE $o_t \gets$ crop$(I_t, g_t)$ \COMMENT{crop stereo foveated vision}
\IF{ATC$(o_t)$ is \textit{global}}
    \STATE $s_{t+1}, s_{t+2}, ..., s_{T} \gets \pi_{global}(o_t, s_t, g_t)$
    \STATE $t' \gets t$
    \WHILE{ATC$(o_t)$ is \textit{global}}
        \STATE Execute $s_{t'}$ on robot
        \STATE$t' \gets t' + 1$
    \ENDWHILE
    \STATE $t \gets t'$
\ELSE 
    \STATE $a_t \gets \pi_{local}(o_t)$ \COMMENT{local action}
    \STATE Execute $a_t$ on robot
    \STATE $t \gets t + 1$
\ENDIF

\STATE $succeed \gets$ subtask success classifier \COMMENT{manual decision or automated (see Section \ref{subsection:consecutive})}
\ENDWHILE
\end{algorithmic}
\end{algorithm}

\subsection{Model architecture}

\begin{figure*}[!t]
\centering
\includegraphics[width=0.8\linewidth]{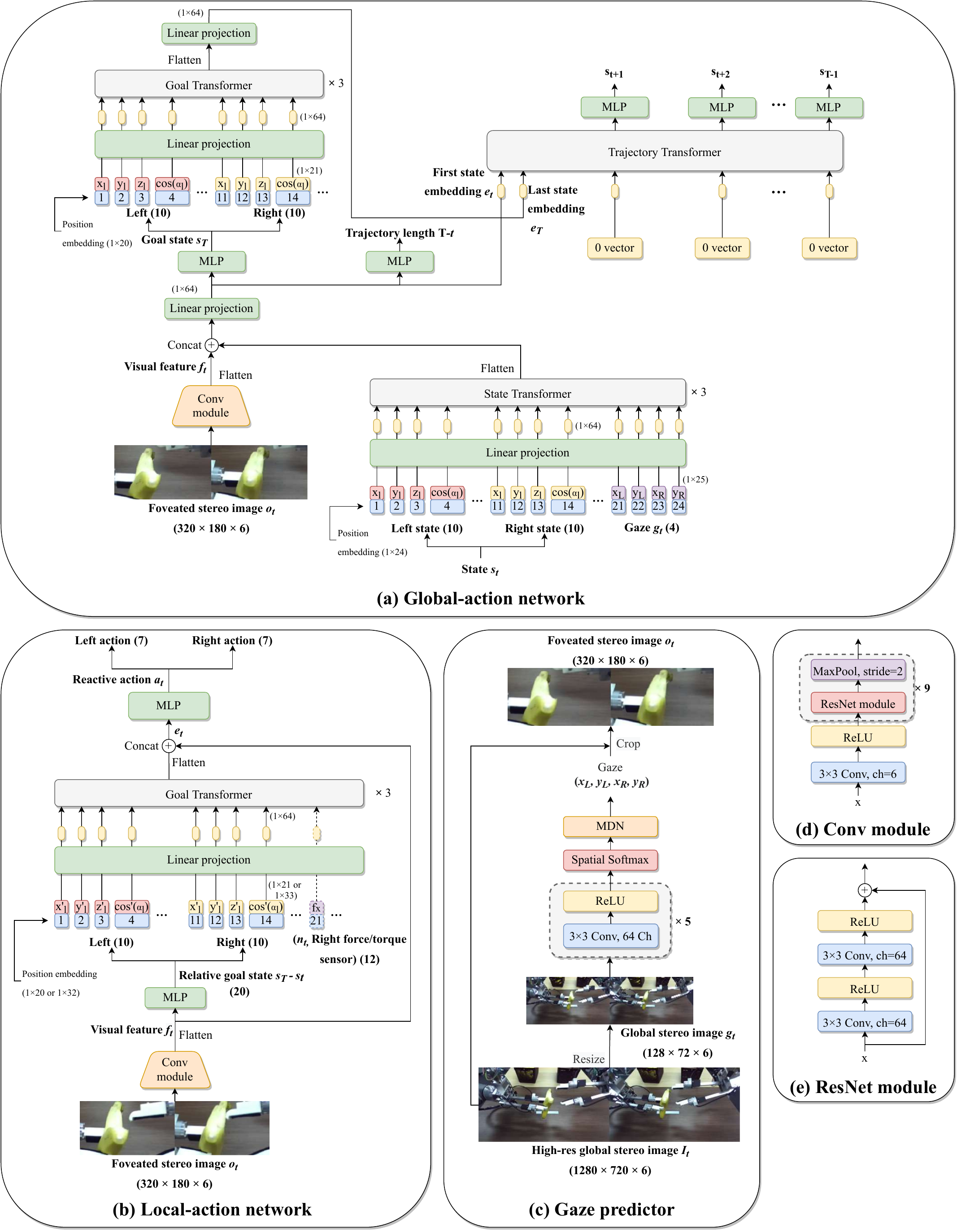}
\caption{Detailed neural network architecture of \textit{GC-DA}. }
\label{fig:network}
\end{figure*}

Figure \ref{fig:network} illustrates the model architecture.
\subsubsection{Gaze predictor}
The gaze predictor uses models similar to those described in \cite{kim2021gaze}. The input $1280 \times 720$ raw image is resized into a $128 \times 72$ global image and then used to predict the gaze coordinate $x_{left}, y_{left}, x_{right}, y_{right}$ measured from the eye tracker while generating the demonstration data. 
Five subsequent convolutional layers and a SpatialSoftmax layer \cite{finn2016deep} process the global image to predict two-dimensional feature coordinates, which then passes through a mixture density network (MDN) \cite{billard2008robot,bazzani2016recurrent} to fit a Gaussian mixture model (GMM) with eight Gaussians into two-dimensional stereo gaze coordinates using the following loss:
\begin{equation}
\label{eq:loss_gaze}
\mathcal{L}_{gaze} = -log \big{(}\sum_{i=1}^{N'} w_t^i \mathcal{N} (h_t;\mu_t^i,\mathbf{\Sigma}_t^i)\big{),}
\end{equation}
where $\mu_t$, $\Sigma_t$, and $w_t$ respectively indicate the mean, covariance matrix, and the weight of each Gaussian, and $h_t$ indicates the measured human gaze during the demonstration. The MDN used two fully connected (FC) layers with rectified linear units (ReLU) between layers. 

During the test, the robot chooses stereo gaze coordinates that maximize the probability of the GMM. The stereo gaze coordinates are used to crop (in stereo) the RGB foveated image ($320 \times 180 \times 6$) from the raw stereo image ($1280 \times 720 \times 6$). 

\subsubsection{Local-action network}\label{subsubsection:local_action}
Local action outputs reactive action using the foveated image. The input foveated image $o_t$ is processed using residual connections \cite{he2016deep} to extract the visual feature $f_t$ using convolutional layers. From this feature, the relative goal states $s^{local}_t = s_T - s_t$ are predicted using the MLP. 

Based on the results obtained from \cite{kim2021transformer}, which demonstrated that the Transformer-based self-attention mechanism can reduce distractions in high-dimensional somatosensory information, the relative goal state is processed using the Transformer encoder architecture (Goal Transformer) to focus on more important features. The output of the Goal Transformer is concatenated with $f_t$, and then processed by the MLP to output left/right reactive action. For \textit{Peel}$\ast$ subtasks, as discussed in \cite{kim2022training}, the right hand's force/torque sensor value is also an input to the Goal Transformer so that the robot can also consider the tactile information during peeling. 

\subsubsection{Global-action network}
The global-action network also inputs $o_t$ to extract visual features $f_t$ using the same ResNet-based architecture described in \ref{subsubsection:local_action}. At this time, because the end-effector is usually out of the field of foveated vision, the left/right robot state and the gaze position are processed using the Transformer encoder (State Transformer) to minimize distraction. Here, $e_t$, which is a linearly projected concatenation of $f_t$ and the State Transformer output, passes the MLP to predict the goal left/right state and a length of trajectory $T-t$. Feature $e_t$ is considered to be the initial embedding of the entire sequence, and the predicted goal state is processed by the Goal Transformer and linearly projected to create the last embedding. $T-t$ zero embeddings are generated to assign space for expected trajectory $s_{t+1}, s_{t+2}, ..., s_{T-1}$. All embeddings are gathered and processed by another Transformer encoder (Trajectory Transformer) to predict trajectory $s_{t+1}, s_{t+2}, ..., s_{T-1}$. The robot executes $s_{t+1}, s_{t+2}, ..., s_{T-1}$ and the inferred goal state $s_T$ sequentially during $ATC(o_{t' \in \{t+1, ..., T \}}) = global$. 

In every neural network architecture, the Transformer that encodes the robot states, gaze, or force/torque sensory input uses a learned position embedding that passes a linear projection because its length is fixed. In addition, the Trajectory Transformer, which generates trajectory action, uses sinusoidal positional embedding \cite{vaswani2017attention} because it has to process the flexible length of the trajectory.

The MLP consists of three FC layers with ReLU between them and 200 nodes in each layer. The training minimizes the $\ell_2$ loss between the model output and the ground-truth action. Other model training details are described in Appendix \ref{appendix:model_train}. Every subtask described in \ref{subsection:task_specification} is trained with a separate model; the subtasks do not share model parameters.

\subsection{Global/local action separation}\label{subsection:global_local_separation}
The proposed dual-action-based method is composed of global and local actions. As mentioned in \ref{dual-action}, this dual-action-based method requires separate training of the local action with the local-action network without disturbances from the global action. For example, in a demonstration sequence with timesteps $\{0, 1, 2, ..., T\}$, 
the steps between $[t_1, t_2]$ are annotated as local action steps, where $t_1$ and $t_2$ are determined from a criterion that decides whether to use local action. The entire set of timesteps $\{0, 1, 2, ..., T\}$ are considered to be the global action.

This subsection explains the criteria used to distinguish global/local actions for the \textit{GraspNeck/Reach}$\ast$ and \textit{Peel}$\ast$ subtasks, which are based on a basic criterion that selects reactive action when precise manipulation is required. Note that this action separation does not change the definition of the goal state; even after they are separated, the global and local actions share the same goal state $s_T$ (e.g., in \textit{GraspReach}$\ast$, goal states are the states where the end-effector reaches the target peel).

\subsubsection{\textit{GraspNeck/Reach}$\ast$ subtasks} 
A step is annotated as a local action step in the  \textit{GraspNeck/Reach}$\ast$ subtasks if the end-effector is in the foveated image (Fig. \ref{fig:boundary}) because, when approaching the target, precise manipulation is required at the end of the sequence, which must perform the last-inch reaching to the exact location of the target \cite{paillard1996fast,kim2021gaze}. Because the \textit{GraspNeck/Reach}$\ast$ subtasks contain many demonstrations ($> 2140$, Table \ref{tab:dataset_statistics}), a CNN-based classifier trained with a small amount of human-annotated data is used to annotate the entire dataset. The details of the automated annotation method can be found in Appendix \ref{appendix:annotation}. The action-type classifier (ATC) is trained based on the annotated labels if the end-effector is in the foveated vision.

\subsubsection{\textit{Peel}$\ast$ subtasks} 
In \textit{PeelNeck}, the criterion is whether the neck of the banana is peeled (Appendix \ref{appendix:global_local}), because a force feedback-based high-precision reactive control is required to tear the neck without severely damaging the banana. After the peel is torn, no force feedback or highly precise reaching to the goal is required. All demonstrations were annotated manually, as the dataset is comparatively small (256 for training and 23 for validation). The annotated data are also used to train another ATC.

In \textit{PeelRight} and \textit{PeelLeft}, the criterion is relatively simple; a step is classified as a global-action step if the end-effector is closed. In these subtasks, the robot must control its end-effector precisely using reactive action during closing so that after complete closure, the end-effector holds the peel. By contrast, the peeling behavior after the holding requires the simple behavior of pulling the firmly held peel, which does not require high-precision manipulation.

\section{Evaluation}
\subsection{Experiment setup}
The banana-peeling task consists of nine subtasks from \textit{GraspBanana} to \textit{PeelLeft}, and each subtask requires the previous subtask to be successful. If all subtasks are conducted consecutively, we can test only a few samples on later subtasks such as \textit{ReachLeft} or \textit{PeelLeft}. Therefore, we tested each subtask by executing the subtask, then modifying the robot and banana position so that the robot could execute the next subtask. For example, if the robot dropped off the banana, it was placed in the robot's hand again. If the end-effector of the robot failed to reach the peel during \textit{ReachRight}, a human manually re-adjusted the robot's end-effector so that it reached the peel. Therefore, even if the robot failed at the previous subtask, the next subtask trial could be evaluated under the same conditions as those used for the trials with successful previous subtasks.

To maintain the condition of the bananas, we bought bananas at the local market and used them on the same day. Each banana was cut to a depth of $1$ cm because even the human operator could not peel the banana through teleoperation because its neck was too stiff to break. We found that the banana became peelable through teleoperation after it had ripened for 2 days. However, the banana also became soft and easily breakable, and controlling the banana's condition became difficult.

\subsection{Ablation studies}
This study evaluated the following questions: 
\begin{itemize}
    \item Is the proposed dual-action system valid?
    \item Is goal-conditioned action inference effective?
\end{itemize}

\begin{table*}[]
\centering
\caption{Ablation study details. }
\begin{tabular}{lcc|ccc}
\hlineB{2}
                & \multicolumn{2}{c|}{Local}           & \multicolumn{3}{c}{Global}                                                                                             \\
Method         & Action type & Goal conditioning & Action type                                                                      & Goal conditioning & Vision type \\\hline \hline
(a) \textit{GC-DA}        & Reactive    & \cmark                & Trajectory                                                                       & \cmark                & Foveated    \\
(b) \textit{NoGC (Global)}    & Reactive    & \cmark                & Trajectory                                                                       & \xmark                & Foveated    \\
(c) \textit{NoGC (Local)}  & Reactive    & \xmark                & Trajectory                                                                       & \cmark                & Foveated    \\
(d) \textit{NoGC (Both)}          & Reactive    & \xmark                & Trajectory                                                                       & \xmark                & Foveated    \\
(e) \textit{Reactive}        & Reactive    & \cmark                & Reactive                                                                         & \cmark                & Foveated    \\
(f) \textit{Reactive-NoGC} & Reactive    & \xmark                & Reactive                                                                         & \xmark                & Foveated    \\
(g) \textit{Traj}   & Trajectory  & \cmark                & Trajectory                                                                       & \cmark                & Foveated    \\
(h) \textit{Dual-resolution} & Reactive    & \cmark                & Trajectory                                                                       & \cmark                & Global    \\ 
\hlineB{2}\\
\end{tabular}
\label{tab:ablation_studies}
\end{table*}

The ablation studies presented in Table \ref{tab:ablation_studies} were designed to solve the above questions. Method (a) \textit{GC-DA} fully utilizes the proposed dual-action architecture and goal-conditioned action inference. The \textit{NoGC} methods ((b)--(d)) do not use the goal-conditioned action inference for global action (b), local action (c), or both (d). Method (e) \textit{Reactive} examines the performance of reactive global action. Method (f) \textit{Reactive-NoGC} combines methods (d) \textit{NoGC (both)} and (e) \textit{Reactive}; global/local actions are not conditioned by the goal state and are reactive. In \textit{Reactive} and \textit{Reactive-NoGC}, global actions are computed from foveated vision, robot state, and gaze. In contrast to method (e), in method (g) \textit{Traj}, both global and local actions are trajectories. In this method, the local action adopts the network architecture of the trajectory-based global-action network, but only inputs the foveated vision. To evaluate the combination of global action with the global vision proposed in \cite{kim2021gaze}, method (h) \textit{Dual-resolution} tested a trajectory-based global action inferred from the global image. In method (h), the global image was processed by five convolutional layers combined with a SpatialSoftmax layer \cite{kim2020using,kim2021gaze}. (i) \textit{Diffusion policy} \cite{chi2023diffusionpolicy}, a recently proposed method that generates visuomotor policy using a denoising diffusion process, is tested to determine whether multimodality residing in human-generated demonstration data can be addressed through the conditional denoising diffusion process. The network architecture employs the CNN-based model described in \cite{chi2023diffusionpolicy}, utilizing the foveated stereo image, gaze, and left/right state as input. This is to maintain consistency with other tested architectures and to enhance robustness against visual distractions in the background. The same hyper-parameters as in \cite{chi2023diffusionpolicy} were used. We note that, since \textit{Diffusion policy} generates an action trajectory, it is equivalent to the trajectory-based action rather than to reactive action.

\subsection{Experiment result}
\begin{table*}[]
\centering
\caption{Results of the ablation studies. G and L refer to Global-action and Local-action network types represented in Table \ref{tab:subtask_details}.}
\begin{tabular}{lccccccc|c}
\hlineB{2}

Method    & \textit{GraspBanana}  & \textit{GraspNeck} & \textit{PeelNeck} & \textit{ReachRight} & \textit{PeelRight} & \textit{ReachLeft} & \textit{PeelLeft} & Mean\\
& G & G $\rightarrow$ L & L $\rightarrow$ G & G $\rightarrow$ L & L $\rightarrow$ G & G $\rightarrow$ L & L $\rightarrow$ G &
\\ \hline \hline
(a) \textit{GC-DA}        & \bf{1.000}     & \bf{0.933}    & 0.833   & \bf{0.933}      & 0.750     & \bf{0.846}     & 0.792 & \bf{0.870}   \\
(b) \textit{NoGC (Global)}    & 0.867 & 0.667    & 0.633   & 0.692      & 0.679     & 0.533     & \bf{0.833} & 0.701   \\
(c) \textit{NoGC (Local)}  & \bf{1.000}     & 0.867    & 0.600   & 0.667      & \bf{0.893}     & 0.462     & 0.667 & 0.736   \\
(d) \textit{NoGC (Both)}          & \bf{1.000}     & 0.800    & 0.200   & 0.467      & 0.643     & 0.800     & 0.731 & 0.663   \\
(e) \textit{Reactive}        & 0.867 & 0.400    & 0.800   & 0.583      & 0.708     & 0.700     & 0.778 & 0.691   \\
(f) \textit{Reactive-NoGC} & \bf{1.000}     & 0.600    & \bf{0.929}   & 0.600      & 0.667     & 0.692     & 0.792 & 0.754   \\
(g) \textit{Traj}   & \bf{1.000}     & 0.000        & 0.833   & 0.000          & 0.571     & 0.231     & N/A  & 0.439 \\  
(h) \textit{Dual-resolution} & \bf{1.000}     & 0.600    & 0.700   & 0.500      & 0.775     & 0.714     & 0.607 & 0.696   \\
(i) \textit{Diffusion policy}\cite{chi2023diffusionpolicy} & 0.889 & 0.111 & N/A & N/A & N/A & N/A & N/A & N/A \\
\hlineB{2}\\

\end{tabular}
\label{tab:accuracy}
\end{table*}

Each ablation study method was tested with 15 bananas. Details of the evaluation criteria are described in Appendix \ref{appendix:experiment}. 

Table \ref{tab:accuracy} presents the experiment results. \textit{PickUp} and \textit{Reposition} were not evaluated because these subtasks are intermediate behaviors that connect the previous subtask to the next subtask; therefore, an evaluation metric would be ambiguous. The proposed \textit{GC-DA} recorded the highest mean score. Remarkably, \textit{GC-DA} scored the highest of all the \textit{GraspNeck/Reach}$\ast$ subtasks. Some ablation methods without the goal-conditioning (methods (b) and (c)) recorded higher scores in \textit{PeelRight} and \textit{PeelLeft}, respectively, but all ablation methods without \textit{GC} ((b)--(d)) recorded lower accuracy in the \textit{GraspNeck/Reach}$\ast$ subtasks. This result infers that goal-conditioned action inference improves manipulation accuracy during precise reaching. By contrast, inferring the exact goal position is not essential for the \textit{Peel}$\ast$ subtask. 

When the global-action network output is reactive instead of a trajectory (methods (e)--(f)) and the local action predicted a trajectory (method (g)), performance dropped in the \textit{GraspNeck/Reach}$\ast$ subtasks. These results also show that the combination of reactive local action and trajectory-based global action is essential.

The result for method (h) \textit{Dual-resolution} indicates that the dual-resolution architecture, in which the global action is computed from the global vision, degrades the reaching performance. In our previous research, we adopted a dual-resolution architecture \cite{kim2021gaze}, and two-dimensional translation of end-effector was required for global action, which was sufficient with the low-resolution global vision. However, in the banana-peeling task, during global action, the robot must calculate a robust and exact three-dimensional trajectory to reach the vicinity of the banana peels, which requires high-resolution foveated vision for depth computation and the suppression of background distractions \cite{kim2020using}.

The results for (i) \textit{Diffusion policy} indicate that it was successful for \textit{GraspBanana} but failed with \textit{GraspNeck}. This suggests that the \textit{Diffusion policy} might not be suitable for tasks that demand high precision.

Comparing the outcomes of (a) \textit{GC-DA}, (g) \textit{Traj}, and (i) \textit{Diffusion Policy}, it can be concluded that methods based on trajectory output are unsuitable for precise manipulation, thus necessitating reactive local action. These results underscore the importance of distinguishing between and learning trajectory-based global and reactive local actions separately.

\begin{figure}[!t]
\centering
\includegraphics[width=0.98\linewidth]{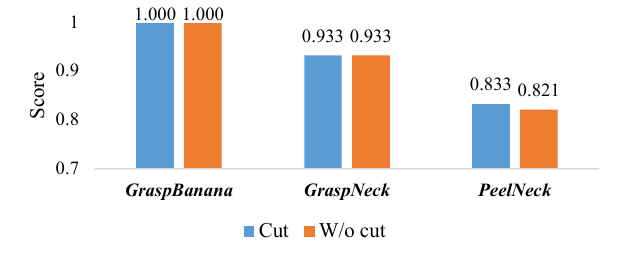}
\caption{Scores up to the \textit{PeelNeck} subtask for experiments using the cut bananas (blue) and the two-day ripened bananas without the cut (orange). The result indicates that there is no difference between the two.}
\label{fig:cut}
\end{figure}

Figure \ref{fig:cut} presents the manipulation score up to the \textit{PeelNeck} subtask for both cut bananas (Table \ref{tab:accuracy} (a)) and bananas ripened for 2 days without the cut. Both showed similar scores, indicating that the banana can be peeled without cutting with enough ripening.

\subsection{Trajectory inference analysis}\label{sec:attention_trajectory}

\begin{figure}[!t]
\centering
\subfloat[Attention map of trajectory-based global action.]{\includegraphics[width=0.8\linewidth]{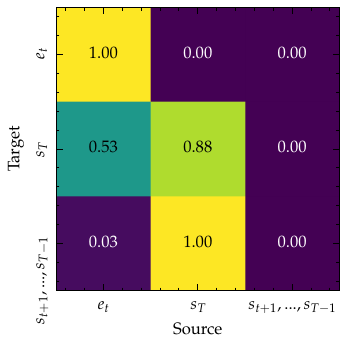}
\label{fig:att_map}}
\hfil
\centering
\subfloat[Attention map of reactive global action.]{\includegraphics[width=0.95\linewidth]{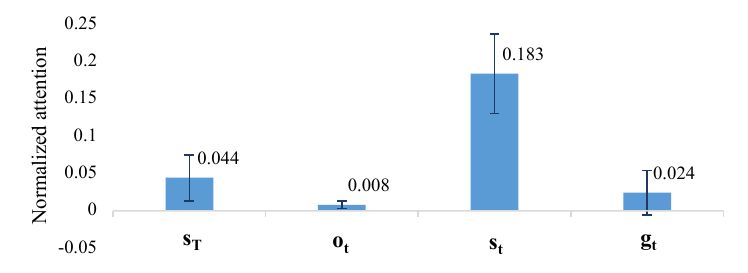}
\label{fig:reactive_attention_map}}
\captionsetup{justification=centering}
\caption{Attention maps of trajectory-based and reactive global actions. The trajectory $s_{t+1}, ..., s_{T-1}$ attended to  goal state $s_T$ (Fig. \ref{fig:att_map}), whereas reactive global action attended more to initial state $s_t$ (\ref{fig:reactive_attention_map}).}\label{fig:attention_maps}
\end{figure}

This subsection examines how trajectory-based global action outperforms reactive global action. The enhanced performance is attributed to the trajectory-based global action being more influenced by the goal state compared to its reactive counterpart, thereby resulting in a more stable policy. For all validation sets across all subtasks, attention maps were computed for both trajectory-based global action (Table \ref{tab:accuracy} (a)) and reactive global action (Fig. \ref{fig:attention_maps}). 

These attention maps were computed by averaging the attention rollout \cite{abnar2020quantifying} of the Transformer's attention weights in embeddings $s_t$, $s_T$, and $[s_{t+1}, ..., s_{T-1}]$ individually. The result clearly shows that the generated trajectories (target: $[s_{t+1}, ..., s_{T-1}]$) are heavily conditioned on the inferred goal state (source: $s_T$, Fig \ref{fig:att_map}). By contrast, Fig. \ref{fig:reactive_attention_map} indicates that the reactive global action is heavily conditioned on current robot kinematic state $s_t$ and less on goal state $s_T$ (see Appendix \ref{appendix:attention_map_cal} for details of the attention computation).

\subsection{Analysis of the goal conditioning}
\begin{figure}[!t]
\centering
\includegraphics[width=0.98\linewidth]{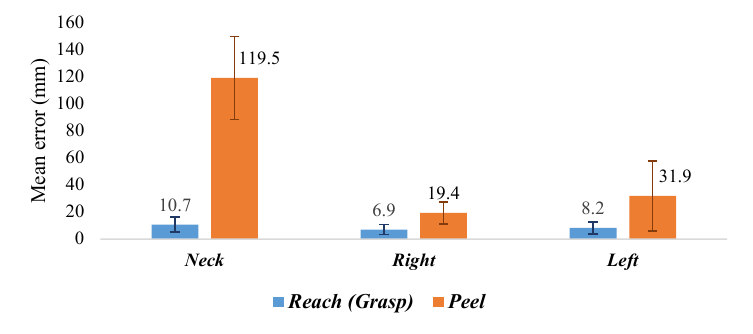}
\caption{Error comparisons of goal state prediction in reach and peel subtasks obtained by global-action networks.}
\label{fig:error_comp}
\end{figure}

This subsection analyzes the typical necessity of goal conditioning during reach subtasks.
Figure \ref{fig:error_comp} visualizes the accuracy of goal state inference on the validation set. In this figure, the mean Euclidean distance between the ground-truth goal state and the predicted goal state of the right arm computed by the global-action network for every timestep is shown. The Euclidean error is low ($<11$ mm) for reach subtasks (blue) but high for peeling subtasks (orange). This result indicates that peeling subtasks are less conditioned by goal state; i.e., peeling can be ended anywhere after the peel is torn, but reaching needs to be ended at the exact location at the banana's peel or neck. 
The above result supports the result in Table \ref{tab:accuracy} (b) and (c), where non-goal-conditioned (\textit{NoGC}) models sometimes showed higher peeling scores than \textit{GC-DA}, as peeling subtasks require less goal conditioned action inference.
Moreover, even though a goal prediction accuracy of $<11$ mm is enough for many pick-and-place tasks, it is not sufficient for reaching the peel in our task setup; therefore, recursive last-inch reaching is required.

\begin{figure}[!t]
\centering
\includegraphics[width=0.98\linewidth]{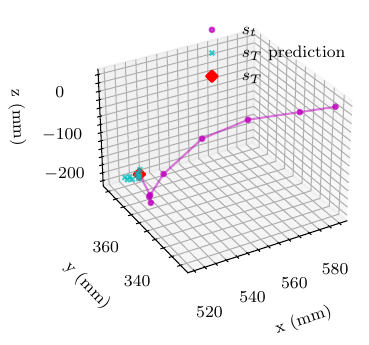}
\caption{Example recorded trajectory and the predicted goal states in \textit{GraspNeck} visualized. State $s_T$ prediction indicates the predicted goal state $s_T$ from every state $s_t$.
}
\label{fig:last_pred}
\end{figure}

Figure \ref{fig:last_pred} visualizes a recorded trajectory $s_0, ..., s_T$ in an episode in the validation set of \textit{GraspNeck} and the predicted goal state $s_T$ from each state $s_t$, $t \in [0, T)$. The goal state can be predicted regardless of current state $s_t$.

\subsection{Consecutive banana peeling}\label{subsection:consecutive}
In previous sections, \textit{GC-DA} was tested on each subtask respectively.
However, in a real-world scenario, the robot must peel the entire banana with all subtasks in order, without human intervention. 
One problem in this scenario is that the final success rate may be dramatically low because failure accumulates throughout the subtasks; the success of the previous subtask is usually required for the success of the current subtask. 
Another problem is that the initial posture and states of the banana cannot be manually adjusted at the start of each subtask because no human intervention is permitted. Therefore, the robot may have to start a subtask even though the banana is in a state in which the robot cannot accomplish the subtask. To solve these problems, it is necessary to (1) maintain a success rate that is as high as possible for each subtask and (2) maintain a stable grasping posture of the banana while peeling. To this end, we implemented a few modifications.

\subsubsection{Modifications for consecutive banana peeling}
\begin{figure*}[!t]
\centering
\includegraphics[width=0.98\linewidth]{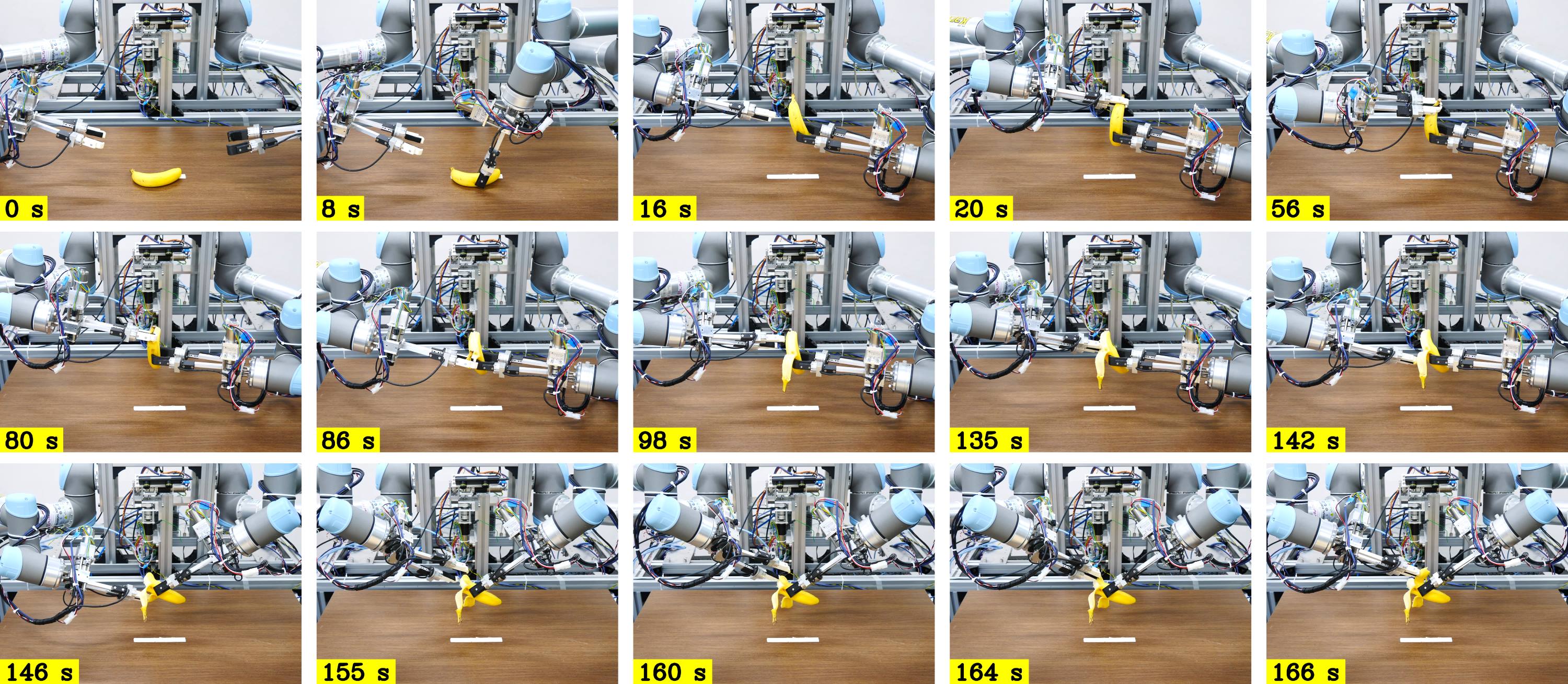}
\caption{
Example of a successful trial. The robot grasped and picked up the banana ($0$--$16$ s), reached its right end-effector to the neck ($16$--$20$ s), then peeled the neck ($20$--$86$ s). Then, the robot reached its right end-effector to the right side of the banana and peeled it ($86$--$142$ s). Finally, the robot rotated the banana ($146$ s) and peeled the left part ($146$--$166$ s).
}
\label{fig:successful}
\end{figure*}

First, the gripper of the left arm was re-designed to be concave so that the robot can grasp the banana more stably. This modification reduces slipping of the banana while it is being peeled (see the left end-effector in Fig. \ref{fig:successful}).

Second, a pre-defined behavior that pushes the banana's neck $30$ mm to the left between \textit{GraspNeck} and \textit{PeelNeck} was adopted to prevent the banana from tilting. Because of the simplicity of this behavior, the pushing action was manually designed.

\begin{figure}[!t]
\centering
\subfloat[\textit{ReachRight-FromRight}.]{\includegraphics[width=0.45\linewidth]{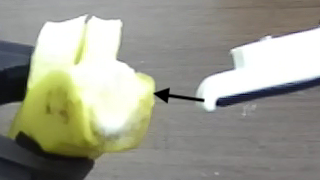}\label{fig:reach1Aug1}}
\subfloat[\textit{ReachRight-FromLeft}.]{\includegraphics[width=0.45\linewidth]{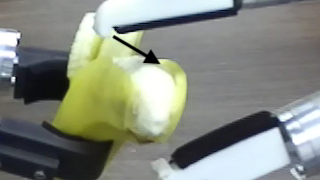}\label{fig:reach1Aug2}}\hfil
\subfloat[\textit{ReachLeft-FromRight}.]{\includegraphics[width=0.45\linewidth]{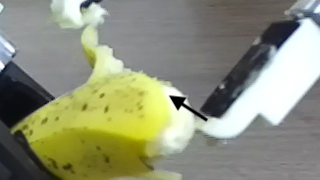}\label{fig:reach2Aug1}}
\subfloat[\textit{ReachLeft-FromLeft}.]{\includegraphics[width=0.45\linewidth]{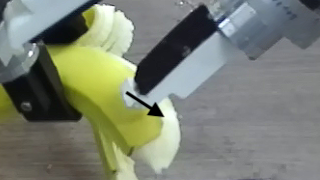}\label{fig:reach2Aug2}}\hfil
\captionsetup{justification=centering}
\caption{Hard-to-reach cases for \textit{ReachRight} and \textit{ReachLeft}.}\label{fig:aug_directions}
\end{figure}

Third, we found that additional reaching data from hard-to-reach cases were helpful. Figure \ref{fig:aug_directions} illustrates examples of such cases; for example, Fig. \ref{fig:reach2Aug2} illustrates an additional demonstration data sample that assumes that the end-effector missed the target peel and is placed on the left side of the peel during \textit{ReachLeft}. The end-effector is required to move back to the peel using a policy that is not usually acquired during a normal demonstration, so an intentionally generated dataset of such state helps the robot to handle failure cases. For subtasks \textit{ReachLeft} and \textit{ReachRight}, we defined two hard-to-reach cases (\textit{FromRight} and \textit{FromLeft} in each subtask (Fig. \ref{fig:aug_directions}), then generated the demonstrations. 

These demonstrations were used in two ways. First, they were integrated into the dataset of Table \ref{tab:dataset_statistics} to increase the number of total demonstrations used for training. Second, each case was separately trained with another local-action network that takes charge when the robot faces each hard-to-reach case.
A simple CNN-based classifier is trained with the foveated images to classify which case the current state is in: the two hard-to-reach cases (\textit{FromLeft} or \textit{FromRight}),  or the normal cases. The training dataset statistics can be found in Table \ref{tab:dataset_statistics_augment}. During the test, the classifier determines which local-action network to use.
Finally, we added additional demonstration sets as well as the demonstrations for the hard-to-reach cases. These demonstrations were used to train the model with the demonstration sets in Table \ref{tab:dataset_statistics}.
Training dataset statistics for the additional demonstrations can be found in Table \ref{tab:dataset_statistics_augment}. 

\begin{table}
\centering
\begin{tabular}{lcc}
\hlineB{2}
    Dataset                   & Number of demos & Total demo time (min) \\ \hline \hline
\textit{GraspBanana}                  & 1429     & 76.7 \\
\textit{PickUp} & 868     & 29.0 \\
\textit{GraspNeck} & 1359     & 55.5 \\ 
\textit{PeelNeck} & 149     & 18.5 \\ 
\textit{ReachRight} (total) & 4233     & 184.6 \\ \hdashline
\multicolumn{1}{r}{(normal)} & \multicolumn{1}{r}{2029}     & \multicolumn{1}{r}{90.3} \\ 
\multicolumn{1}{r}{\textit{FromRight}} & \multicolumn{1}{r}{1174}     & \multicolumn{1}{r}{51.1} \\ 
\multicolumn{1}{r}{\textit{FromLeft}} & \multicolumn{1}{r}{1030}     & \multicolumn{1}{r}{40.2} \\ \hdashline
\textit{PeelRight} & 156     & 21.3 \\ 
\textit{Reposition} & 158     & 4.37 \\ 
\textit{ReachLeft} (total) & 3355     & 159.8 \\ \hdashline
\multicolumn{1}{r}{(normal)} & \multicolumn{1}{r}{1284}     & \multicolumn{1}{r}{74.4} \\ 
\multicolumn{1}{r}{\textit{FromRight}} & \multicolumn{1}{r}{1125}     & \multicolumn{1}{r}{52.0} \\ 
\multicolumn{1}{r}{\textit{FromLeft}} & \multicolumn{1}{r}{946}     & \multicolumn{1}{r}{33.3} \\ \hdashline
\textit{PeelLeft} & 146     & 12.3 \\ \hline 
\textit{Total} & 11853 & 561.8 \\
\hlineB{2}\\

\end{tabular}
\caption{Additional training set statistics.}
\label{tab:dataset_statistics_augment}
\end{table}

\subsubsection{Transition between subtasks}
There are two different types of transitions between adjacent subtasks. First, when the earlier subtask ends with trajectory inference (\textit{GraspBanana}, \textit{PickUp}, and \textit{Reposition}), and the \textit{Peel}$\ast$ subtasks, later subtasks can be conducted as soon as the current trajectory execution is done. Otherwise, if the subtask ends with a reactive action (\textit{GraspNeck/Reach}$\ast$ subtasks), the transition between subtasks requires additional classifiers. 

The robot changes \textit{GraspNeck} to \textit{PeelNeck} when the right end-effector is closed. \textit{ReachLeft} and \textit{ReachRight} use additional CNN-based classifiers (the NSC). In \textit{ReachRight}, the NSC is trained with classified foveated images in \textit{ReachRight} and \textit{PeelRight}; if the current foveated image $o_t$ is classified as \textit{PeelRight}, the next subtask for execution is changed to \textit{PeelRight}. In \textit{ReachLeft}, another NSC classifies foveated images in \textit{ReachLeft} and \textit{PeelLeft}. The NSC comprises the convolution module described in Fig. \ref{fig:network} (d) and a three-layer MLP. 

\subsubsection{Result}
\begin{figure}[!t]
\centering
\subfloat[
The robot failed to reach its end-effector to the peel.
]
{\includegraphics[width=0.98\linewidth]{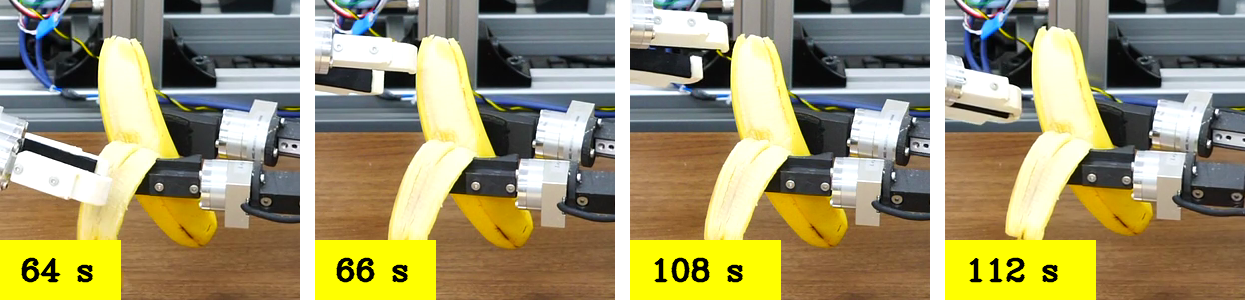}
\label{fig:fail0}}
 \vspace{8mm}
\subfloat[
The banana's neck was broken during peeling.
]{\includegraphics[width=0.98\linewidth]{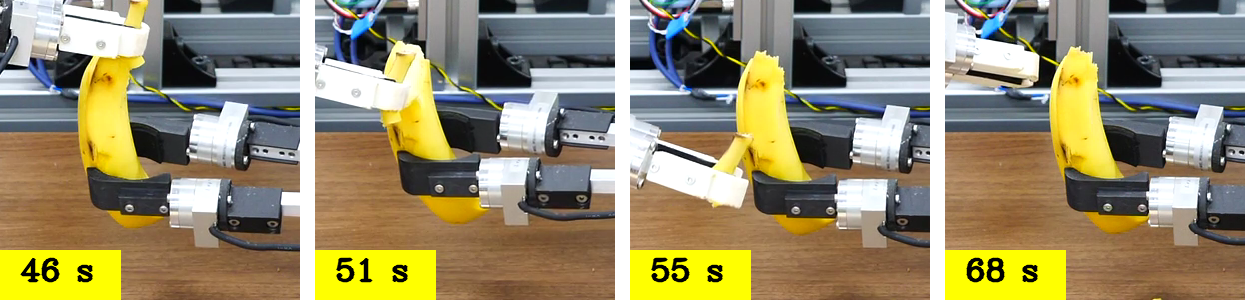}
\label{fig:fail1}}
\captionsetup{justification=centering}
\caption{Failure examples of banana peeling.}\label{fig:fail}
\end{figure}

Figure \ref{fig:successful} illustrates an example of the successful behavior of \textit{GC-DA} in the consecutive banana peeling setup. By contrast, Fig. \ref{fig:fail} shows failure examples for the same setup, including examples in which the robot failed to reach the end-effector to the target (Fig. \ref{fig:fail0}) or the neck was broken during peeling (Fig. \ref{fig:fail1}). 

\begin{figure}[!t]
\centering
\includegraphics[width=0.98\linewidth]{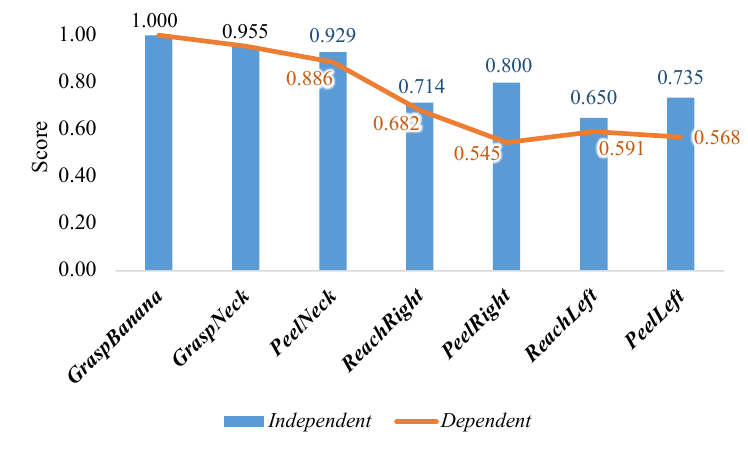}
\caption{Score of each subtask in consecutive banana peeling. \textit{Independent} refers to the score of each subtask; therefore it does not consider failures from previous subtasks, whereas \textit{Dependent} also counts failures from the previous subtask.
}
\label{fig:acc_score}
\end{figure}

Figure \ref{fig:acc_score} presents the score of consecutive banana peeling with 22 bananas. \textit{Independent} considers only the executed trials for each subtask: \textit{Independent} $= \sum_i^N {score}_i / N$ for $N <= 22$ is the number of trials actually executed in each subtask. For example, if $21/22$ trials succeeded to perform \textit{GraspNeck}, the $N$ for \textit{PeelNeck} is $21$. \textit{Dependent} also considers unexecuted trials: \textit{Dependent} $= \sum_i^{22} {score}_i / {22}$. Note that the \textit{Dependent} score for \textit{ReachLeft} can be higher than the \textit{Dependent} score for \textit{PeelRight} because the robot can attempt the \textit{ReachLeft} subtask even though it has failed at the  \textit{PeelRight} subtask.

Table \ref{tab:success_rate_consecutive} reports each peeling subtask's success rate and the overall subtask success rate. The success rate is calculated for trials with \textit{scores} $> 0.5$.

\begin{table}
\centering
\caption{Success rate for each peeling task and total success rate}
\begin{tabular}{lcccc}
\hlineB{2}
Subtask & \textit{PeelNeck} & \textit{PeelRight} & \textit{PeelLeft} & \textit{Total} \\ \hline \hline
Success rate (\%) & 95.45 & 59.09 & 63.63 & 40.91 \\
\hlineB{2} \\ 
\end{tabular}
\label{tab:success_rate_consecutive}
\end{table}

\begin{table}
\centering
\caption{Task success rate comparison of \textit{GC-DA}, \textit{Reactive-NoGC}, and \textit{Diffusion policy} on \textit{MoveBowl} and \textit{PickCoin}.}
\begin{tabular}{lccc}
\hlineB{2}
Task & \textit{GC-DA}  & \textit{Reactive-NoGC} & \textit{Diffusion policy}\\ \hline \hline
\textit{MoveBowl} & \textbf{93.3\%} & 0.0\% & 6.7\% \\
\textit{PickCoin} & \textbf{70.0\%} & 30.0\% & 5.0\%\\
\hlineB{2} \\ 
\end{tabular}
\label{tab:comparison_multi}
\end{table}

\begin{figure}[!t]
\centering
\subfloat[
\textit{MoveBowl}: robot supports right and left side of the bowl with its both arms and transport it to the tray.
]
{\includegraphics[width=0.98\linewidth]{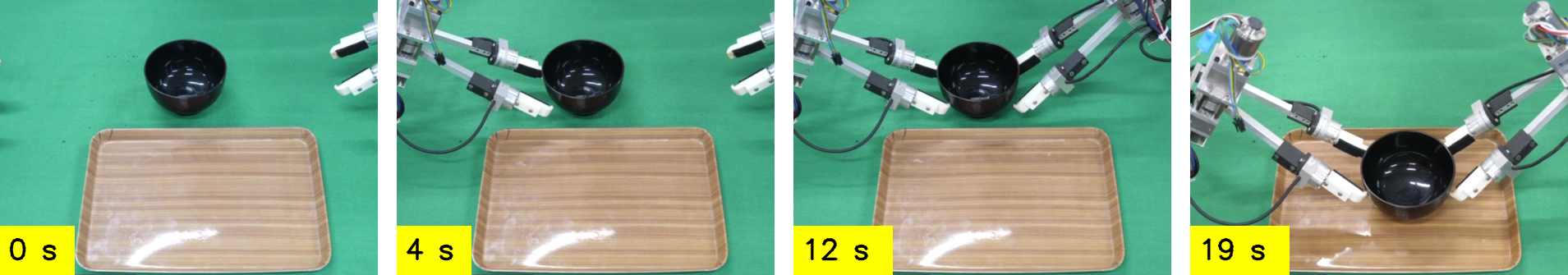}
\label{fig:fail0_}}
 \vspace{8mm}
\hfil
\subfloat[
\textit{PickCoin}: robot picks coin on the table.
]{\includegraphics[width=0.98\linewidth]{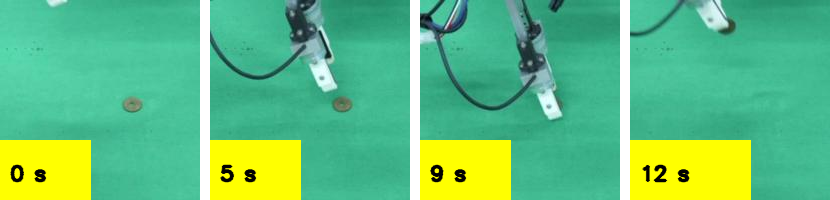}
\label{fig:fail1_}}
\captionsetup{justification=centering}
\caption{
Example successful trials of \textit{MoveBowl} and \textit{PickCoin} with \textit{GC-DA}.}\label{fig:mutli_task}
\end{figure}

\begin{figure}[!t]
\centering
\subfloat[Configurations with varying combinations of known and novel objects and backgrounds.]{\includegraphics[width=0.98\linewidth]{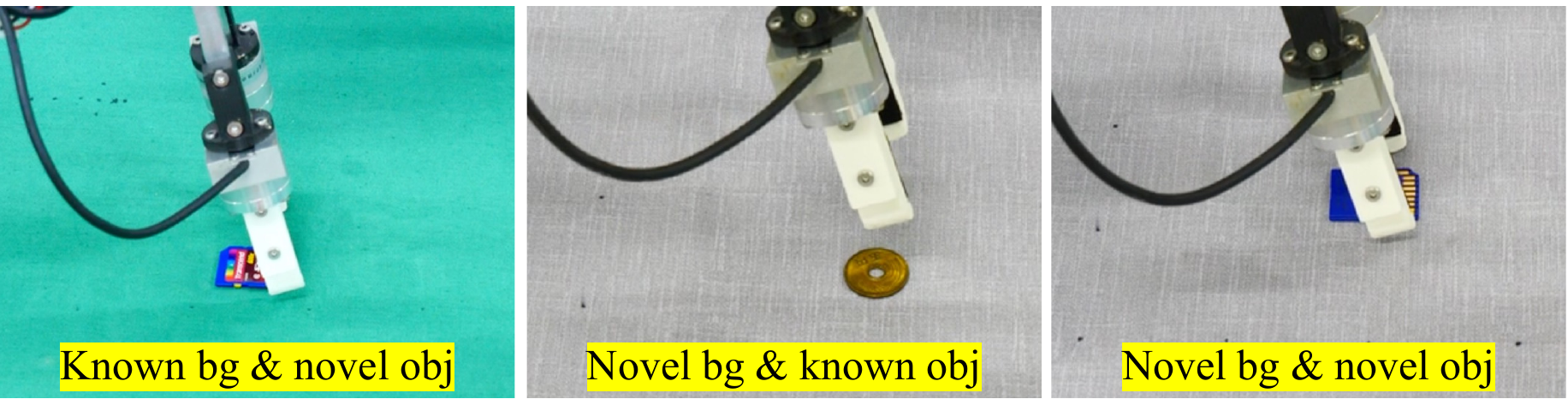}\label{fig:config_pick}}\hfil
\subfloat[Success rates.]{\includegraphics[width=0.98\linewidth]{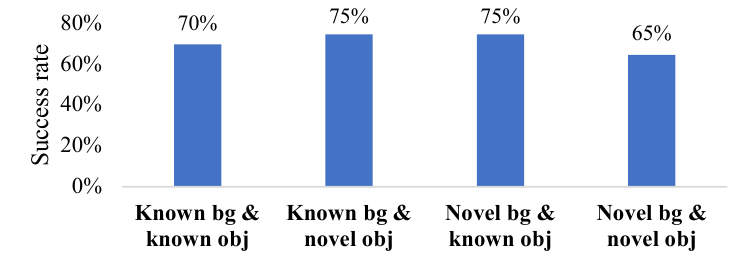}\label{fig:robust_pick}}\hfil
\captionsetup{justification=centering}
\caption{Robustness of \textit{GC-DA} with various combinations of known and novel objects (obj) and backgrounds (bg).}\label{fig:result_robust_pick}
\end{figure}

\subsection{Applicability of \textit{GC-DA}}
To demonstrate the applicability of the proposed method to other manipulation tasks, we trained and tested the model on two additional tasks: \textit{MoveBowl} and \textit{PickCoin}. As illustrated in Fig. \ref{fig:mutli_task}, in \textit{MoveBowl}, the robot supports the bowl from the right and left sides with both arms and move the bowl to the tray. This task requires dual-arm manipulation and balancing of the bowl while moving. It is divided into three subtasks: moving the right end-effector to the right side of the bowl (\textit{SupportRight}), moving the left end-effector to the left side of the bowl (\textit{SupportLeft}), and moving the bowl to the tray with both arms (\textit{Move}). In \textit{PickCoin}, the robot picks up a thin coin from the table, which requires precise control.
Table \ref{tab:dataset_statistics_multi} presents the training set statistics for both tasks.

\begin{table}
\centering
\caption{Training set statistics of \textit{PickCoin} and \textit{MoveBowl}. G and L refer to global-action and local-action network types.}
\begin{tabular}{lcc}
\hlineB{2}
Dataset & Number of demos & Total demo time (min) \\ \hline \hline
\textit{PickCoin (G $\rightarrow$ L)} & 3051 & 168.7 \\
\textit{MoveBowl} (total) & 2430 & 84.7 \\ \hdashline
\multicolumn{1}{r}{\textit{SupportRight} (G $\rightarrow$ L)} & \multicolumn{1}{r}{840} & \multicolumn{1}{r}{24.4} \\
\multicolumn{1}{r}{\textit{SupportLeft} (G $\rightarrow$ L)} & \multicolumn{1}{r}{712} & \multicolumn{1}{r}{30.4} \\
\multicolumn{1}{r}{\textit{Move} (G)} & \multicolumn{1}{r}{878} & \multicolumn{1}{r}{29.9} \\
\hlineB{2}
\end{tabular}
\label{tab:dataset_statistics_multi}
\end{table}

Table \ref{tab:comparison_multi} presents the results for both tasks. In each case, the final success rate was significantly enhanced by using the proposed \textit{GC-DA}. \textit{Diffusion policy} reached the vicinity of the target object but failed at precise object manipulation (e.g., picking up a coin, or accurately supporting the bowl with both arms), indicating the necessity of reactive local action for dexterous, high-precision manipulation.
Additionally, to demonstrate the learned policy's robustness to novel objects or backgrounds, further experiments were conducted with \textit{PickCoin} using a novel object (a memory card) and/or a novel background (gray) (Fig. \ref{fig:config_pick}). The results indicate that the proposed model can successfully adapt to a novel object and/or background (Fig. \ref{fig:robust_pick}).

\subsection{Robustness of gaze prediction}
\begin{figure}[!t]
\centering
\subfloat[Error by number of visual distractors.]{\includegraphics[width=0.98\linewidth]{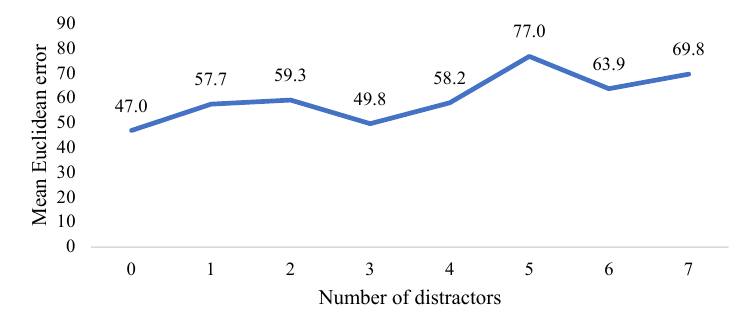}\label{fig:error_gaze}}\hfil

\subfloat[Error across different backgrounds with seven distractors.]{\includegraphics[width=0.98\linewidth]{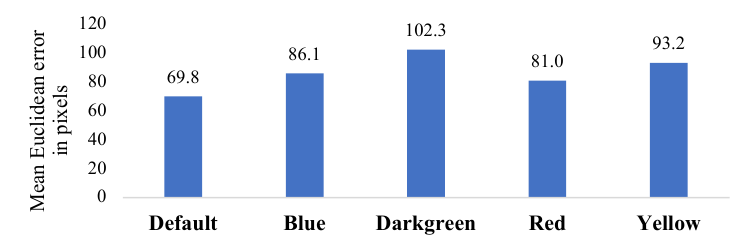}\label{fig:novel_bg_obj}}\hfil
\captionsetup{justification=centering}
\caption{Mean Euclidean distance error in pixels of the gaze prediction against the number of visual distractors in the background and background variability.}\label{fig:background_gaze_res}
\end{figure}

\begin{figure}[!t]
\centering
\subfloat[\textit{GraspBanana}]{\includegraphics[width=0.48\linewidth]{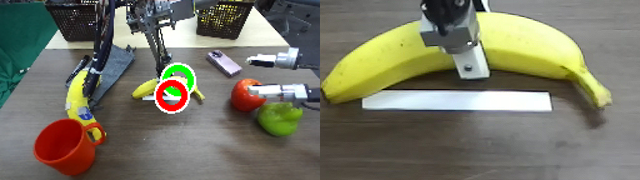}
\label{fig:gaze_0}}
\subfloat[\textit{GraspNeck}]{\includegraphics[width=0.48\linewidth]{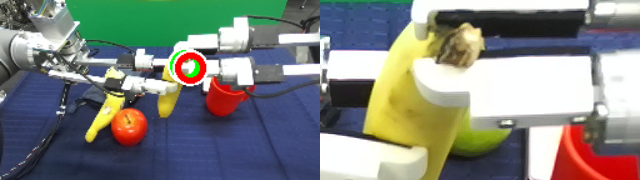}
\label{fig:gaze_2}}  
\hfil
\subfloat[\textit{PeelNeck}]{\includegraphics[width=0.48\linewidth]{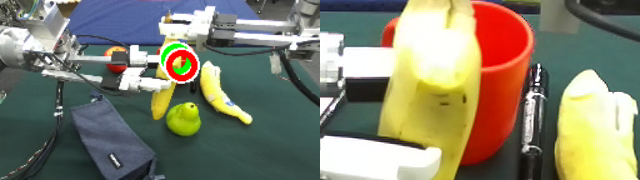}
\label{fig:gaze_3}}
\subfloat[\textit{ReachRight}]{\includegraphics[width=0.48\linewidth]{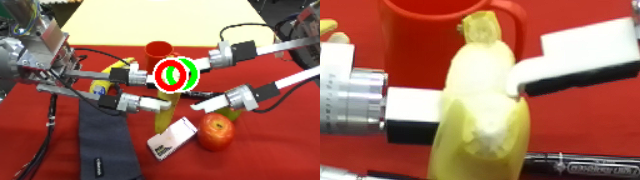}
\label{fig:gaze_4}}
\hfil
\subfloat[\textit{PeelRight}]{\includegraphics[width=0.48\linewidth]{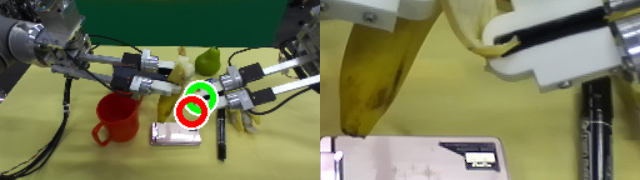}
\label{fig:gaze_5}}
\subfloat[\textit{ReachLeft}]{\includegraphics[width=0.48\linewidth]{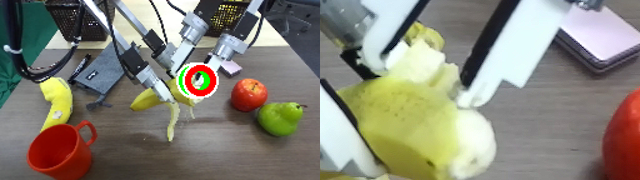}
\label{fig:gaze_7}}
\hfil
\subfloat[\textit{PeelLeft}]{\includegraphics[width=0.48\linewidth]{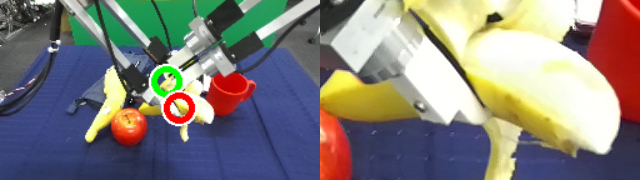}
\label{fig:gaze_8}}

\captionsetup{justification=centering}
\caption{Example gaze prediction results in a cluttered environment (7 distractors with various backgrounds). The left images display the human gaze (green) and predicted gaze (red) on the left camera image, while the right images showcase the foveated vision generated using the predicted gaze. Even with a background cluttered with various objects, the predicted gazes successfully focused on the target while excluding distractors.}\label{fig:gaze_res}
\end{figure}

The robustness of gaze prediction against visual distractors in the background is crucial, as the robot may need to execute object manipulation in novel environments. We evaluated the gaze predictor using a teleoperated banana peeling dataset, incrementally increasing the number of distractors in the background. Figure \ref{fig:error_gaze} presents the mean Euclidean error in pixels with various numbers of distractors, showing an increasing tendency of the error as the number of distractors rises. A slight increase in the error was also observed with various backgrounds (Fig. \ref{fig:novel_bg_obj}). However, as depicted in Fig.  \ref{fig:gaze_res}, even in the most complicated scene (with 7 distractors) and various backgrounds, the predicted gaze successfully attends to the target banana, effectively excluding background distractors.

\section{Conclusion}
This study realized the banana-peeling task with a general-purpose dual-arm robot (UR5) based on the proposed  (\textit{GC-DA}) deep imitation learning. The proposed method separates the policy of robot motion into policies for global and local actions. 
A trajectory-based global action was used to deliver the end-effector stably close to the goal position when the robot does not require very precise manipulation of the target object, such as when delivering the robot arm to the vicinity to the banana peel. The reactive local action precisely controls the end-effector to manipulate the target object, such as during the precise last-inch reaching to the banana peel.
The inferred goal state conditions the model's output action for stable policy inference. 

The proposed \textit{GC-DA} method successfully executed the peeling of real bananas utilizing a dual-arm manipulator. The ablation study, as depicted in Table \ref{tab:ablation_studies}, reveals that the reactive local action was indispensable for achieving manipulation skills with high precision, while the trajectory-based global action proved essential for tasks with a long horizon. Consequently, for tasks requiring precise manipulation over long horizons, a combination of both trajectory-based global action and reactive local action was necessary.

This result is important for two reasons: first, highly dexterous, long-horizon robot manipulation of a deformable object is possible without a task-specific gripper. Second, the above dexterous manipulation skills were acquired from human-generated behavior data.
To this end, the proposed method accomplished the banana-peeling task, which is a combination of many different sub-tasks for grasping the banana, precisely grasping each peel, then peeling, which requires dexterity.

In this research, $811.1 + 561.8$ minutes $\approx$ $21.8$ hours of demonstration data were generated and used for training, which can be obtained in only a few days. Considering the complexity of the target banana-peeling task, we believe that $21.8$ hours of demonstration is feasible. However, data efficiency has always been pursued in learning-based robotics. Therefore, our future interest is to study whether few-shot learning of highly dexterous manipulation skills is possible by transferring knowledge about objects and behaviors. 
Previously, the importance of behavior segmentation and structured behavior and tasks have been reported \cite{Kuniyoshi1994learning,konidaris2012robot,zeng2018semantic}. Because unsupervised segmentation is possible with gaze \cite{yu2002understanding}, and gaze information is highly correlated with task objective \cite{pelz2001coordination,hayhoe2005eye}, we hypothesize that gaze can be applied for the unsupervised segmentation and structuring of tasks and further be applied to the transfer of behavior to other tasks. The application of gaze-based human behavior segmentation for structured deep imitation learning is a promising topic for future research.

This research demonstrated \textit{GC-DA} with the banana-peeling task. Even though the proposed method still requires a few human engineering factors, such as manual task segmentation and dual-action annotation, the proposed method greatly diminishes the amount of manual engineering needed to develop the dexterous manipulation skills required for banana-peeling tasks. Also, our approach relies on a natural semantic segmentation strategy (e.g., "grasp the neck of the banana," "peel the neck of the banana") align well with the semantic structure of the task. Each subtask is defined as a combination of local and global actions which can be easily defined based on the required dexterity (Fig. \ref{fig:task_desc}). In this context, we believe that if users wish to apply our learning-based approach to their own applications, they can perform segmentation in a human-natural manner and allocate local actions according to the necessary dexterity for challenging tasks, such as precise reaching or object manipulation. Therefore, the \textit{GC-DA} schema can also be used for other manipulation tasks that require dexterous skills that cannot be defined by human-engineered rules. For example, our approach is a promising method for various other bimanual manipulation tasks, such as cutting or peeling vegetables (e.g., green onions or onions), where one hand needs to hold the object and the other hand performs the task, and precise location and orientation of the end-effectors are crucial. In such cases, the proposed dual-action can effectively succeed in approaching the end-effector (i.e., knife) to near the target place and actually cut the onion with the reactive local-action. 
Another potential application of the proposed method is factory automation of deformable objects, such as smartphone box packaging, cable manipulation, or food manipulation.
We expect further research will apply the \textit{GC-DA} to those manipulation tasks while reducing the remaining effort needed for manual annotation.

\appendix
\section{Appendix}
\subsection{Performance of ATC and NSC}
This subsection presents the accuracy results of the ATC and NSC on the training and validation sets (Tables \ref{tab:ATC_acc} and \ref{tab:NSC_acc}).

\begin{table}
\centering
\caption{Accuracy of the ATC on the training and validation sets.}
\begin{tabular}{lcc}
\hlineB{2}
Subtask & Training (\%) & Validation (\%) \\ \hline \hline
\textit{GraspNeck} & 94.91 & 96.48 \\
\textit{PeelNeck} & 94.23 & 88.24 \\
\textit{ReachRight} & 93.69 & 93.33 \\
\textit{ReachLeft} & 93.29 & 93.51 \\
\hlineB{2} \\ 
\end{tabular}
\label{tab:ATC_acc}
\end{table}

\begin{table}
\centering
\caption{Accuracy of the NSC on the training and validation sets.}
\begin{tabular}{lcc}
\hlineB{2}
Subtask & Training (\%) & Validation (\%) \\ \hline \hline
\textit{ReachRight} & 97.59 & 94.99 \\
\textit{ReachLeft} & 98.19 & 97.52 \\
\hlineB{2} \\ 
\end{tabular}
\label{tab:NSC_acc}
\end{table}

\subsection{Boundaries between global/local actions in the \textit{GraspNeck/Reach}$\ast$ subtasks}\label{appendix:global_local}
\begin{figure}[!t]
\centering
\subfloat[\textit{GraspNeck}.]{\includegraphics[width=0.98\linewidth]{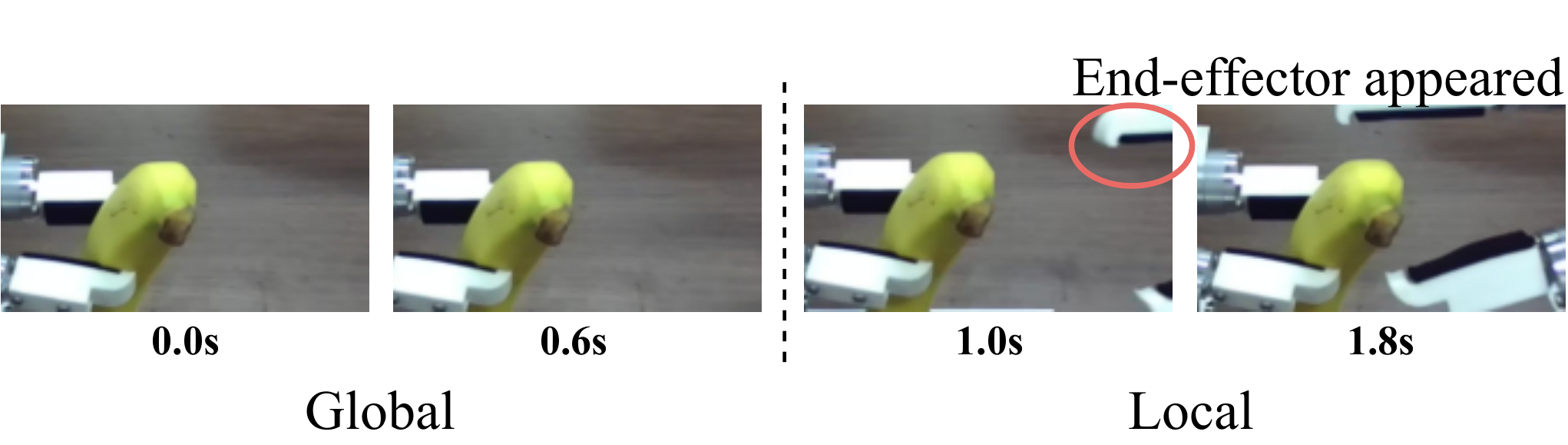}\label{fig:reach0}}\hfil
\subfloat[\textit{PeelNeck}.]{\includegraphics[width=0.98\linewidth]{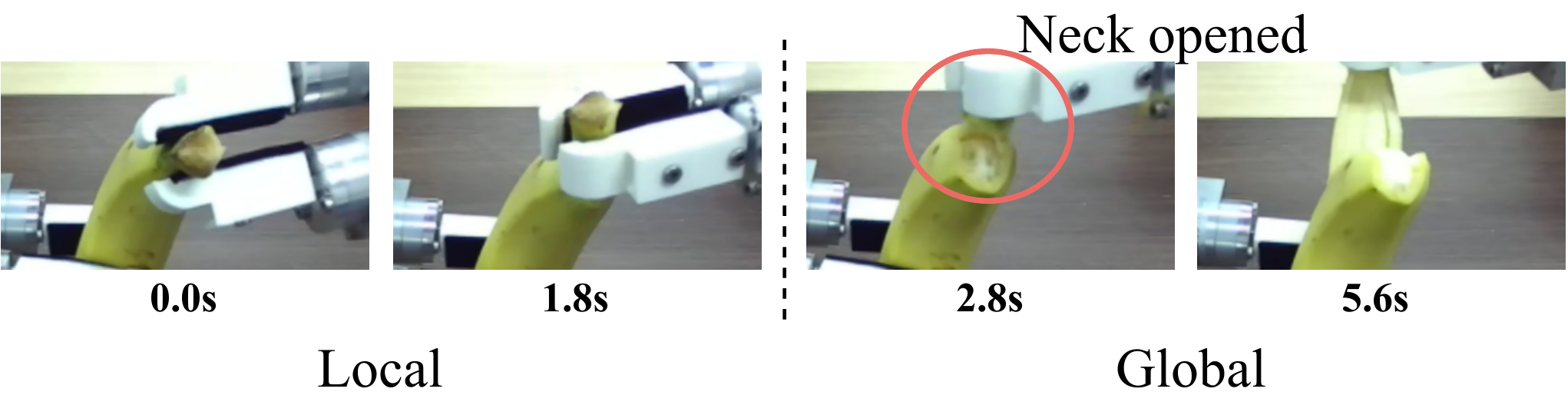}\label{fig:peel0}}\hfil
\subfloat[\textit{ReachRight}. The effective end-effector is the upper end-effector.]{\includegraphics[width=0.98\linewidth]{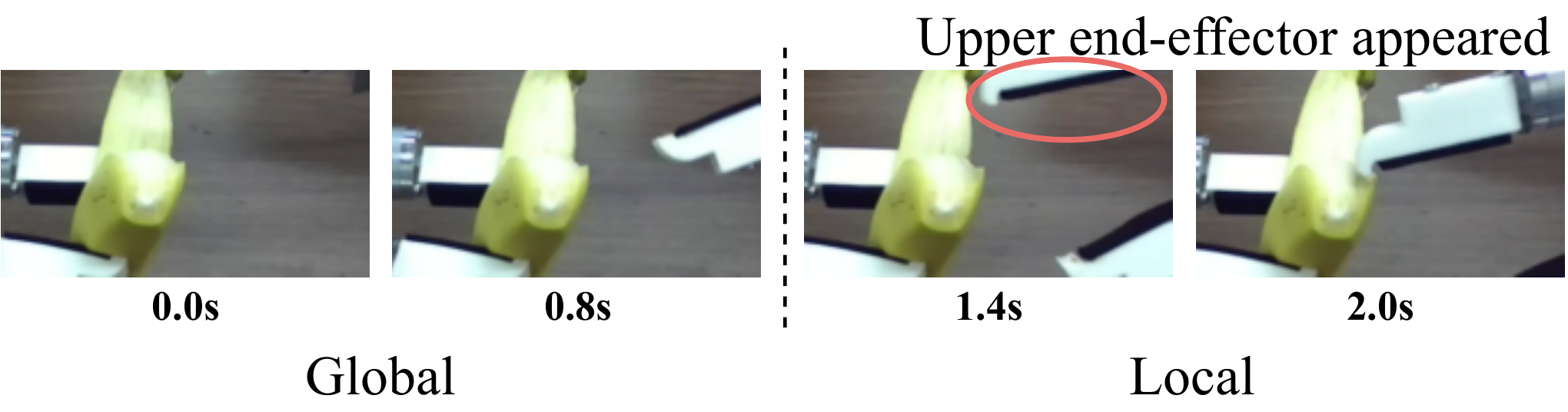}\label{fig:reach1}}\hfil
\subfloat[\textit{PeelRight}.]{\includegraphics[width=0.98\linewidth]{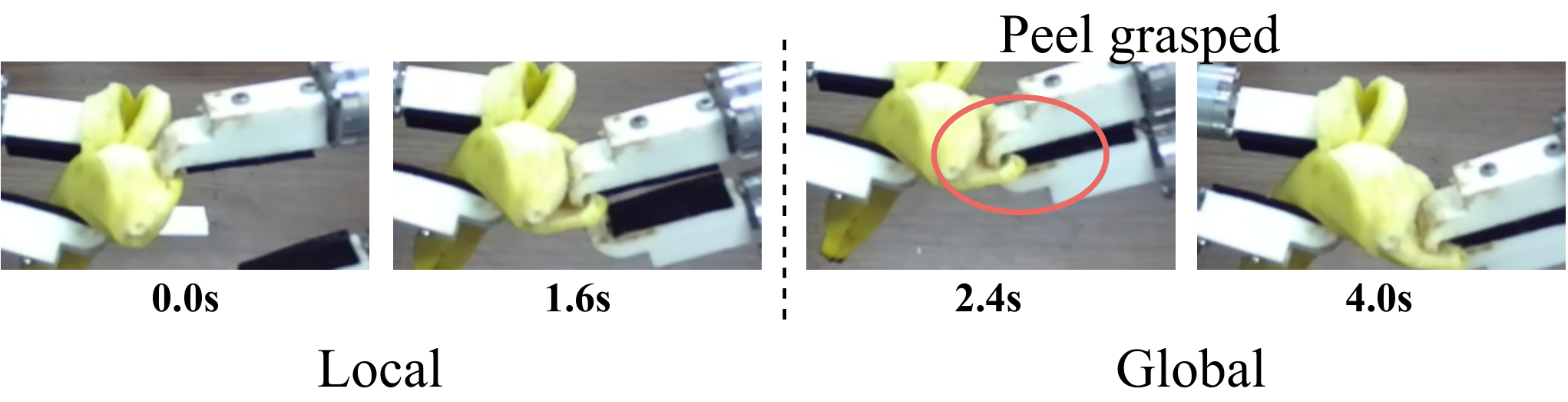}\label{fig:peel1}}\hfil
\subfloat[\textit{ReachLeft}. The effective end-effector is the lower end-effector.]{\includegraphics[width=0.98\linewidth]{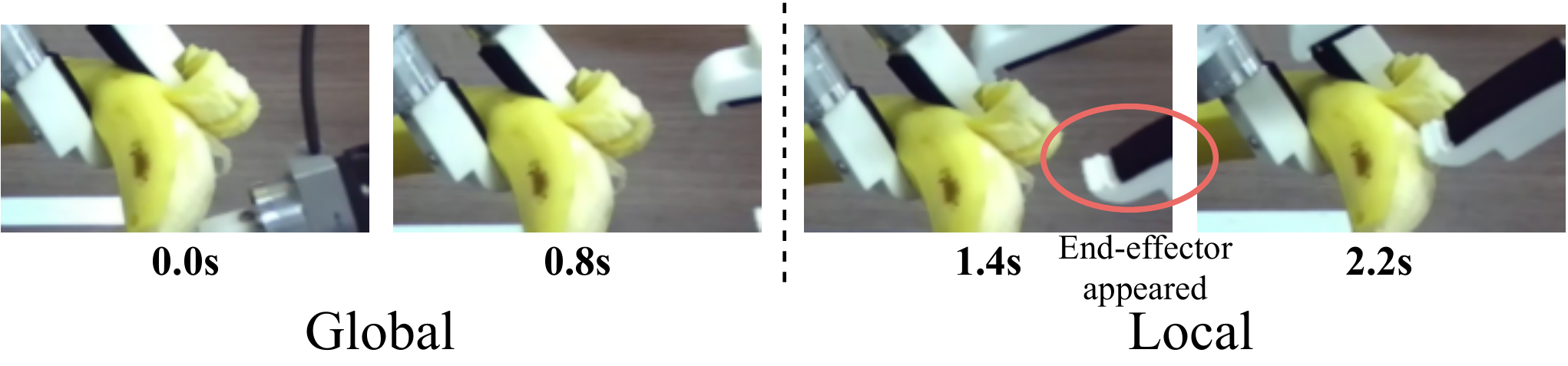}\label{fig:reach2}}\hfil
\subfloat[\textit{PeelLeft}.]{\includegraphics[width=0.98\linewidth]{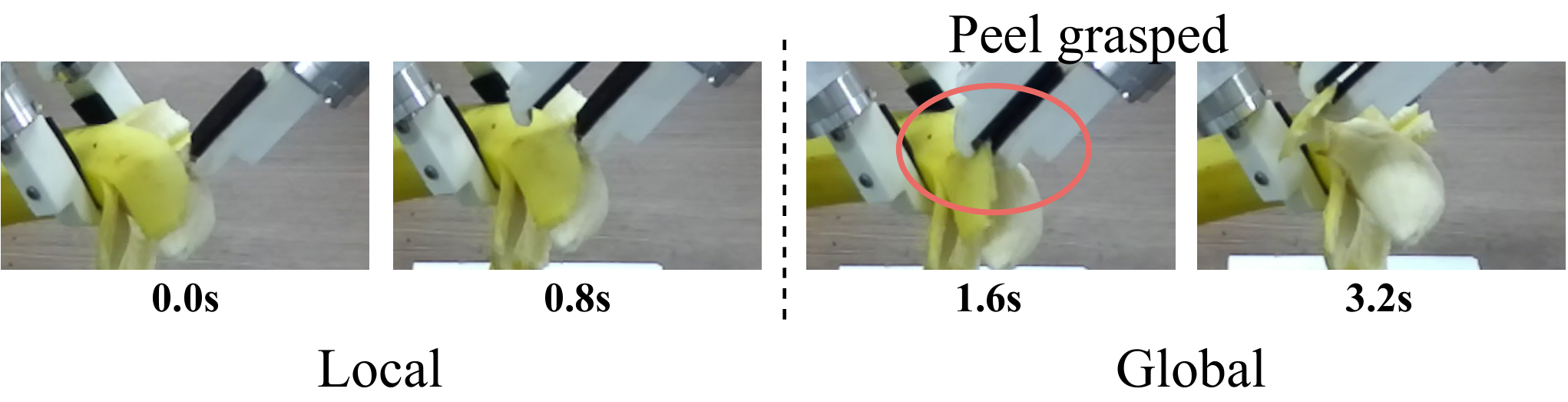}\label{fig:peel2}}\hfil
\captionsetup{justification=centering}
\caption{Boundaries displayed in the foveated image.}\label{fig:boundary}
\end{figure}

Figure \ref{fig:boundary} illustrates examples of boundaries between global and local actions in each subtask. In the \textit{GraspNeck/Reach}$\ast$ subtasks, a step is identified as a local action if the effective end-effector is in the foveated image (Figs. \ref{fig:reach0}, \ref{fig:reach1}, and \ref{fig:reach2}). In \textit{PeelNeck} (Fig. \ref{fig:peel0}), a behavior that opens the banana's neck is identified as a local action, and the peeling behavior is identified as the global action. In \textit{PeelLeft} (Fig. \ref{fig:peel1}) and \textit{PeelRight} (Fig. \ref{fig:peel2}), grasping the banana peel is defined as a local action, and peeling behavior is identified as a global action. 

\subsection{Annotation method} \label{appendix:annotation}
To reduce the time required for annotation, we used an image classifier for automated annotation similar to \cite{kim2022memory}'s. First, a human manually annotated part of the entire episode into global/local actions. For example, a human annotated 283 out of 2370 episodes in the \textit{GraspNeck} subtask. Then, a simple CNN-based image classifier was trained using the annotated dataset (90\% training, 10\% validation). This CNN-based classifier comprised ten consecutive convolutional layers with ReLU, and a maxpooling layer was placed between each convolutional layer. The output of the convolutional layers was passed through two layers of MLPs, which returned the  following binary classification results:
$C(o_t) \rightarrow \{\textit{global action}, \textit{local action}\}$, where $C$ indicates the trained classifier. Next, the trained classifier annotated the unannotated episodes. To filter false classification results, we utilized the fact that the transition between global and local actions occurs only once. In the \textit{GraspNeck/Reach}$\ast$ subtasks, if $C(o_t) \rightarrow \textit{local action}$ and $C(o_{t+1}) \rightarrow \textit{global action}$ for any $t$ in timestep $0, ..., T-1$ in an episode of length $T$, this episode was re-annotated by the human. For the \textit{Peel}$\ast$ subtasks, the human re-annotated the episodes with $C(o_t) \rightarrow \textit{global action}$ and $C(o_{t+1}) \rightarrow \textit{local action}$. Table \ref{tab:annot_count} presents the number of demonstrations annotated by humans that were used for the model training, classified demonstrations by the trained model, and re-annotated demonstrations.

\begin{table}
\centering
\caption{Number of annotated, classified by the trained model, and re-annotated demonstrations on each subtask.}
\begin{tabular}{lccc}
\hlineB{2}
Subtask & \# of annotated & \# of classified & \# of re-annotated \\ \hline \hline
\textit{GraspNeck} & 283 & 2087 & 369 \\
\textit{ReachRight} & 400 & 4546 & 288 \\
\textit{ReachLeft} & 400 & 4065 & 202 \\
\hlineB{2} \\ 
\end{tabular}
\label{tab:annot_count}
\end{table}

\subsection{Model training details}\label{appendix:model_train}
This research followed the model training details described in \cite{kim2022training}. A rectified Adam optimizer \cite{liu2019variance} with a learning rate of $1e-5$ for the global and local-action networks and $1e-4$ for the gaze predictors, a weight decay of $0.01$, and a batch size of 16 were used to train the proposed model. Two AMP EPYC 7402 CPUs with four NVIDIA a100 and Intel Xeon CPU E5-2609 v4 with eight NVIDIA v100 GPUs were used for training the model. Because the \textit{GraspNeck/Reach}$\ast$ subtasks are more complex than the other subtasks, they were trained for 500 epochs, and the other subtasks were trained for 300 epochs. During the execution on the robot, an  Intel CPU Core i7-8700K and one NVIDIA GeForce GTX 1080 Ti were used. 

\subsection{Scoring criteria}\label{appendix:experiment}
The model was given a score of $+1$ if it achieved the objective of each subtask and $+0$ if it did not. In addition, on the \textit{Peel}$\ast$ subtasks, specific scoring criteria were set. Detailed explanations of the criteria are as follows: 
\begin{itemize}
    \item \textit{GraspBanana}: $+1$ if the robot grasps the banana without any harm. $+0$ if the robot destroys the banana or collides with the table.
    \item \textit{GraspNeck}: $+1$ if the robot grasps the banana's neck. $+0$ if the robot fails to grasp the neck.
    \item \textit{PeelNeck}: $+0.5$ for opening the neck and $+1$ for entirely peeling the neck. 
    \item \textit{ReachRight}: $+1$ if the upper end-effector touches the border between the right peel and pulp. $+0$ otherwise.
    \item \textit{PeelRight}: If the bottom side of the peel is still attached, the robot scores $+0.5$. If it is entirely torn, the robot scores $+1$. 
    \item \textit{ReachLeft}: $+1$ if the lower end-effector touches the border of the left peel and pulp. $+0$ otherwise.
    \item \textit{PeelLeft}: If the bottom side of the peel is still attached, the robot scores $+0.5$. If it is entirely torn, the robot scores $+1$. 
\end{itemize}

\begin{figure}[!t]
\centering
\subfloat[\textit{PeelNeck}: score $= 1$.]{\includegraphics[width=0.8\linewidth]{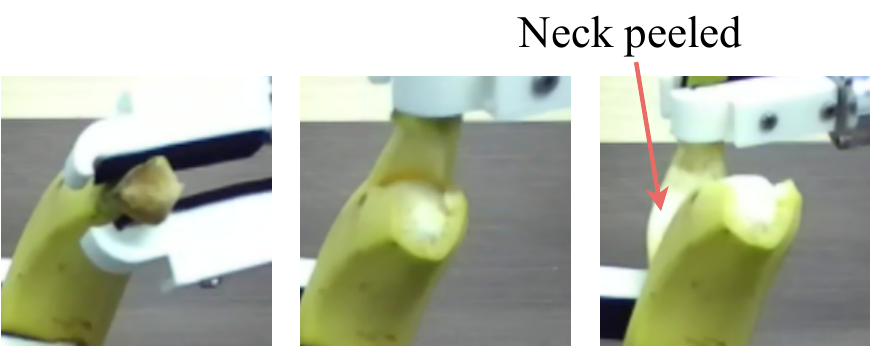}
\label{fig:peel0_1}}
\hfil
\subfloat[\textit{PeelNeck}: score $= 0.5$.]{\includegraphics[width=0.8\linewidth]{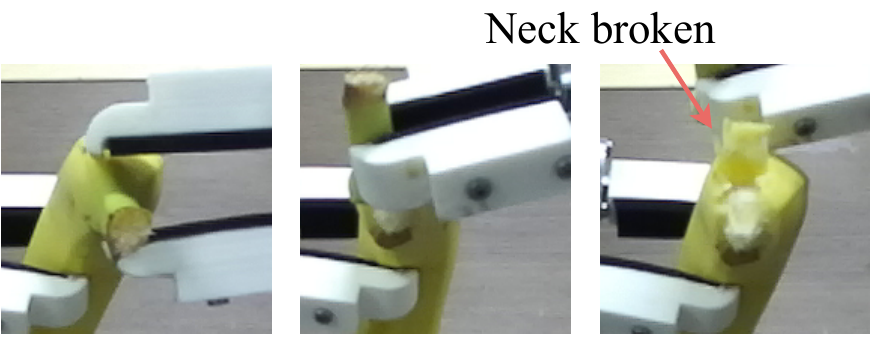}
\label{fig:peel0_05}}  
\hfil
\subfloat[\textit{PeelRight}: score $= 1$.]{\includegraphics[width=0.8\linewidth]{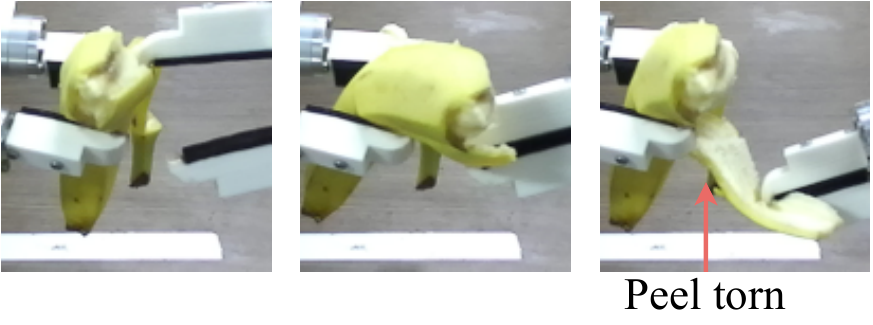}
\label{fig:peel1_1}}
\hfil
\subfloat[\textit{PeelRight}: score $= 0.5$.]{\includegraphics[width=0.8\linewidth]{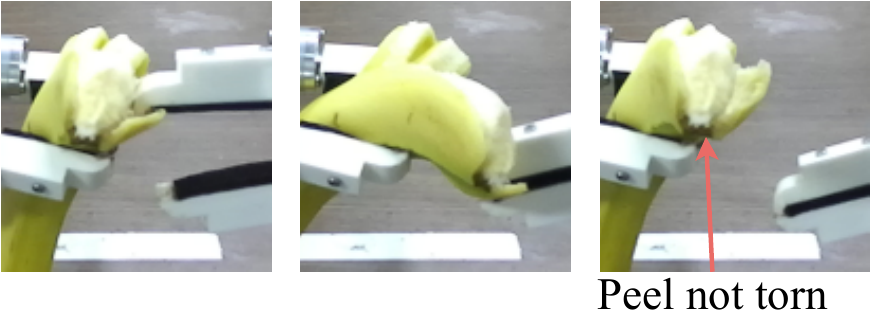}
\label{fig:peel1_05}}  
\hfil
\subfloat[\textit{PeelLeft}: score $= 1$.]{\includegraphics[width=0.8\linewidth]{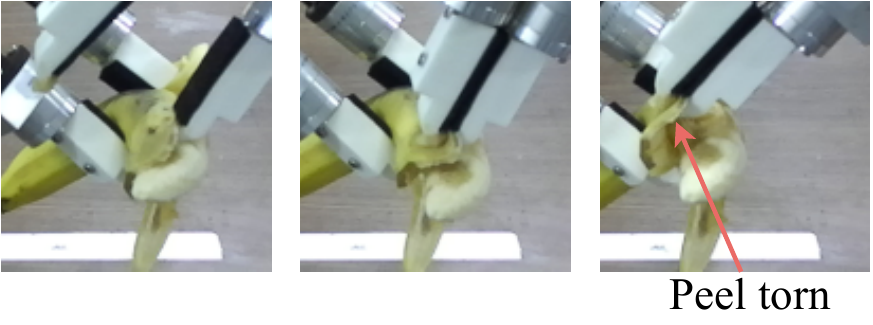}
\label{fig:peel2_1}}
\hfil
\subfloat[\textit{PeelLeft}: score $= 0.5$.]{\includegraphics[width=0.8\linewidth]{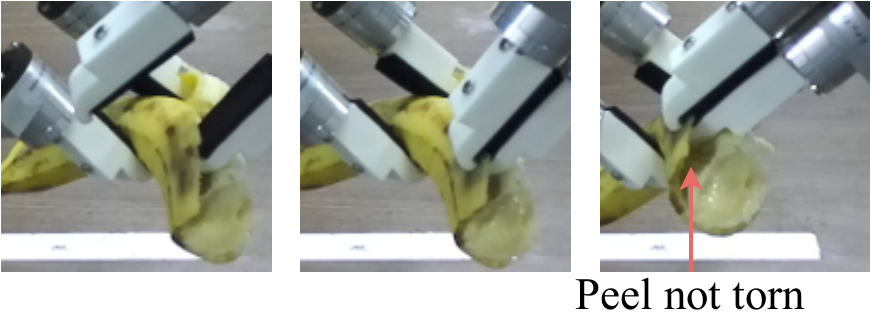}
\label{fig:peel2_05}}
\captionsetup{justification=centering}
\caption{Examples of scoring criteria.}\label{fig:criteria}
\end{figure}

Figure \ref{fig:criteria} visualizes examples of these criteria.

\subsection{Translational and rotational uncertainties in foveated vision}\label{appendix:training_distributions}
\begin{figure*}[!t]
\centering
\subfloat[\textit{GraspNeck}.]{\includegraphics[width=0.95\linewidth]{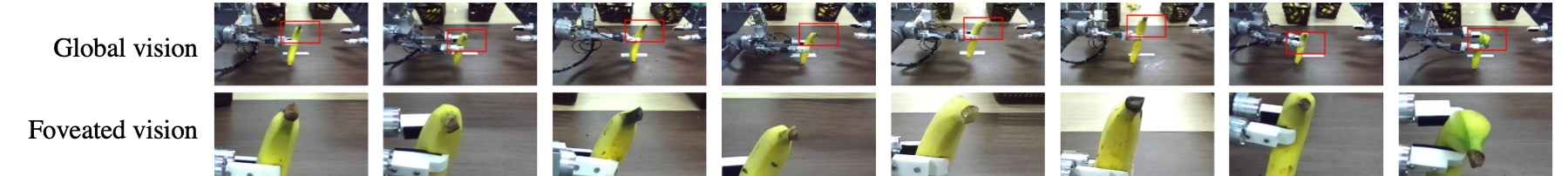}
\label{fig:train_dist_0}}
\hfil
\subfloat[\textit{ReachRight}.]{\includegraphics[width=0.95\linewidth]{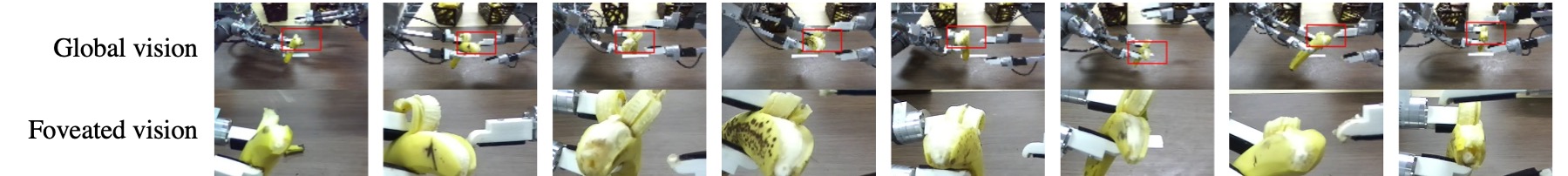}
\label{fig:train_dist_1}}
\hfil
\subfloat[\textit{ReachLeft}.]{\includegraphics[width=0.95\linewidth]{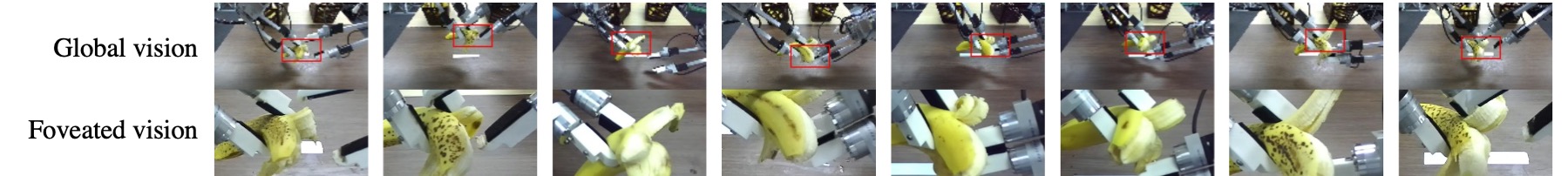}
\label{fig:train_dist_2}}
\captionsetup{justification=centering}
\caption{Samples of the robot's view of subtasks \textit{GraspNeck} (a), \textit{ReachRight} (b), and \textit{ReachLeft} (c). The robot effectively addresses the translational uncertainties of the target object by utilizing gaze, i.e., despite the target object potentially varying in global vision, gaze can effectively capture the target's position so that it always appears in the foveated vision.}\label{fig:training_dist}
\end{figure*}

Figure \ref{fig:training_dist} illustrates a few samples of the initial step of the robot's perspective on subtasks \textit{GraspNeck}, \textit{ReachRight}, and \textit{ReachLeft}. The translational uncertainties of the target object were efficiently managed using gaze, while rotational uncertainties remain in the foveated vision.

\subsection{Data augmentation}
Image augmentation is often used in computer vision to increase the robustness of the neural network to input data. This research randomly adjusted the brightness, contrast, saturation, and hue to transform the training input images. Additionally, random erasing \cite{zhong2020random} was applied to prevent overfitting of the neural network model to the specific features that only exist in the training data.

\subsection{Attention computation}\label{appendix:attention_map_cal}
The computation of the attention map followed \cite{kim2022memory}. The trajectory-based global-action network inputs current state embedding $e_t$, inferred goal state $s_T$, and zero vectors $s_{t+1}, ..., s_{T-1}$. On this input sequence, attention rollout \cite{abnar2020quantifying} $A^{\{T-t+1\} \times \{T-t+1\}}$ defined by the recursive multiplication of the attention weight of each layer in the trajectory Transformer is computed. To normalize the attention rollout on each target embedding, the attention rollout is divided by the maximum of target embedding values as follows:
\begin{equation}
\begin{aligned}
\label{eq:attention}
M_{i,j} = \frac{A_{i,j}}{max(A_{i})},
\end{aligned}
\end{equation}
where $i \in [0, T)$, $j \in [0, T)$ and $M$ is the sequential attention map.
Then, the attention on zero vectors $s_{t+1}, ..., s_{T-1}$ is averaged. The output attention map size is $3 \times 3$, where the elements are denoted as $[e_t, s_T, \{s_t+1,... s_T-1\}]$.

Similarly, Eq. \ref{eq:attention} normalizes an attention rollout of the reactive global-action network output. The Transformer output embeddings of the reactive global-action network are flattened and passed to an MLP to compute $a_t$. Therefore, the information about which source embedding is paid attention is more important than the attention map that considers the target embedding. The summation of target embedding corresponding to $o_t$, $s_t$, $g_t$, and $s_T$ is computed by the following:
\begin{equation}
\begin{aligned}
\label{eq:summed0}
&W^{s_T} = \sum^{138}_{i=0} \sum^{19}_{j=0} A_{ij},\\
\end{aligned}
\end{equation}
\begin{equation}
\begin{aligned}
\label{eq:summed1}
&W^{o_t} = \sum^{138}_{i=0} \sum^{114}_{j=20} A_{ij},\\
\end{aligned}
\end{equation}
\begin{equation}
\begin{aligned}
\label{eq:summed2}
&W^{s_t} = \sum^{138}_{i=0} \sum^{134}_{j=115} A_{ij},\\
\end{aligned}
\end{equation}
\begin{equation}
\begin{aligned}
\label{eq:summed3}
&W^{g_t} = \sum^{138}_{i=0} \sum^{138}_{j=135} A_{ij}.\\
\end{aligned}
\end{equation}

\section*{Acknowledgments}
This paper is partly based on results obtained under a Grant-in-Aid for Scientific Research (A) JP22H00528 and supported in part by the Department of Social Cooperation Program ``Intelligent Mobility Society Design,'' funded by Toyota Central R\&D Labs., Inc., of the Next Generation AI Research Center, The University of Tokyo.

\section{References Section}
 
%
\bibliographystyle{IEEEtran}
\bibliography{IEEEfull}

\newpage

 
\vspace{11pt}

\vfill

\end{document}